\documentclass[11pt]{article}

\usepackage[final]{acl}

\usepackage{times}
\usepackage{latexsym}

\usepackage[T1]{fontenc}

\usepackage[utf8]{inputenc}

\usepackage{microtype}

\usepackage{inconsolata}

\usepackage{graphicx}

\usepackage{amsfonts}
\usepackage{amsmath}
\usepackage{booktabs}

\usepackage[most]{tcolorbox}
\usepackage{fvextra}
\usepackage{xcolor}

\definecolor{promptbg}{RGB}{248,248,248}
\definecolor{promptframe}{RGB}{180,180,180}

\newtcolorbox{promptbox}[1][]{
  enhanced,
  breakable,
  colback=promptbg,
  colframe=promptframe,
  boxrule=0.6pt,
  arc=2mm,
  left=2mm,
  right=2mm,
  top=1mm,
  bottom=1mm,
  fonttitle=\bfseries,
  title=#1
}

\title{Implicit Personality Representations in Humans and LLMs}

\author{
 \textbf{Yilin Geng\textsuperscript{1}},
 \textbf{Omri Abend\textsuperscript{2}},
 \textbf{Eduard Hovy\textsuperscript{1}},
 \textbf{Lea Frermann\textsuperscript{1,3}} \\
 \textsuperscript{1}University of Melbourne,
 \textsuperscript{2}Hebrew University of Jerusalem, 
 \textsuperscript{3}University of Tübingen \\
 \small{
   \textbf{Correspondence:} \href{mailto:yilin.geng@student.unimelb.edu.au}{yilin.geng@student.unimelb.edu.au}
 }
}

\begin{document}
\maketitle
\graphicspath{{./}}
\begin{abstract}

A century of psychology has found that the trait words people use to describe one another vary, but the relational structure among those traits, which ones go together and which oppose, is strikingly consistent across raters and cultures. We test whether the LLM (Qwen 2.5-7B-Instruct) reproduces this structure in its internal trait representations. From millions of crowd-sourced personality ratings of fictional characters, we build a human implicit-personality matrix over hundreds of traits; from contrastive model activations, we build a matching matrix over the same traits. The two relational structures align strongly (Mantel $r = 0.77$), and the agreement holds trait by trait as well as in aggregate. Two dominant axes of the model's trait representations recover the social and intellectual dimensions long known to organize human personality impressions, social warmth and intellectual competence. On held-out dialogue, projecting model activations onto these directions yields personality profiles that agree with human ratings. This work establishes a framework that enables comprehensive, human-grounded comparison between internal model trait geometry and the shared structure of human personality impressions.

\end{abstract}

\section{Introduction}

Humans are remarkably efficient and consistent at assigning a personality to individuals, and they do so by attributing stable characteristics ---  or {\it traits} --- such as brave, stubborn, or curious that are inferred from how a person speaks and acts~\citep{chaplin1988conceptions}. 
A century of research on personality psychology has established structured personality representations~\citep{allport1936traitnames, goldberg1990alternative}, and has shown that while the trait-describing terms vary among individuals and cultures~ \citep{albright1988consensus}, their covariance or {\it structure} of this representational space is highly consistent~\citep{schneider1973implicit,shweder1980systematic}. 

Trained on the text that people produce, large language models (LLMs) exhibit personality-relevant behavior~\citep{serapio2025psychometric, lee2025trait}, and individual concepts can be located as directions in their internal representations~\citep{park2024linear, arditi2024refusal, chen2025persona}. Prior output-level work has also recovered known personality factors from model probabilities over trait descriptors~\citep{suh2024rediscovering}. This paper investigates whether representations (independently extracted internal trait directions) in the LLM activation space reproduce a relational structure measured from human personality impressions, using a comprehensive, human-grounded framework.

\begin{figure}[t]
\centering
\includegraphics[width=\linewidth]{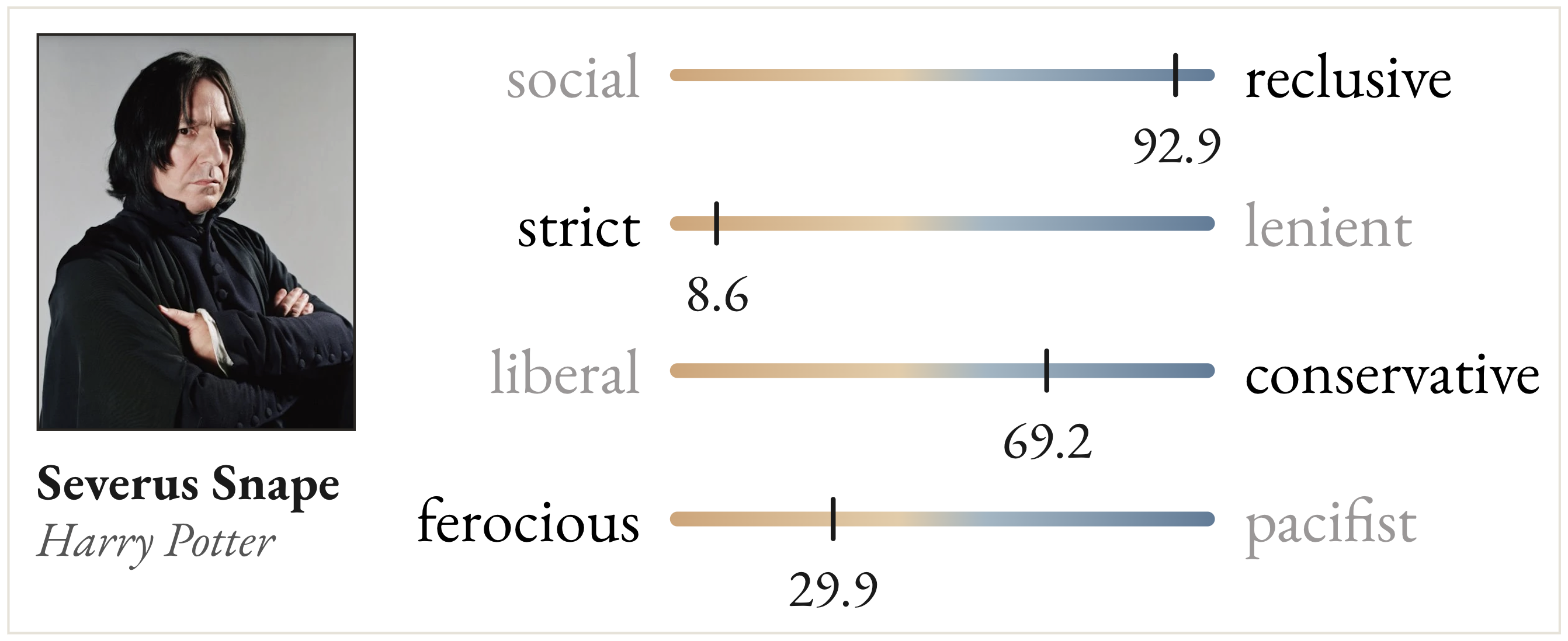}
\caption{Example profile (simplified) of Severus Snape from \textit{Harry Potter}. Each trait scale is a spectrum defined and constrained by two poles of contrastive words or phrases. This work covers 385 such trait scales.}
\label{fig:char_profile_example}
\end{figure}

Our framework builds on a large inventory of over 380 traits, where each trait is represented as a spectrum with two poles (See Figure~\ref{fig:char_profile_example}). The set of traits is derived from a large-scale data set of {\it human} ratings of fictional characters, and we use the traits and their human ratings to ground the LLM representations. Specifically, we use the Statistical ``Which Character'' Personality Questionnaire (SWCPQ), a crowd-sourced data set in which volunteers contributed tens of millions of ratings for thousands of fictional characters~\citep{openpsychometrics2023swcpq}. We thus extend prior work on internal LLM personality representation, which was constrained to a handful of traits which were defined ad-hoc~\citep{arditi2024refusal,chen2025persona}.

We analyse the LLM's internal representation of traits directly, rather than inferring from model output, because it has been shown repeatedly that model behavior does not always accurately reflect its internal mechanisms~\cite{turpin2023language}. Prior work has prompted LLMs to respond to personality questionnaires and has inferred personality such as a ``Big-5 profile'' based on their responses~\citep{serapio2025psychometric, lee2025trait,huang-etal-2024-reliability}, however, this approach does not conclusively answer how LLMs {\it represent} personality, and does not allow us to compare their representational space to the one established in human personality psychology.

Instead, we build on the {\it contrastive prompting} method of \citet{chen2025persona}, improving its clarity, scalability, and validation process. Given a trait with two extreme poles (Figure~\ref{fig:char_profile_example}), we prompt the LLM with scenarios that would elicit divergent behavior by characters corresponding to the respective poles (see an example in Figure~\ref{fig:extraction_pipeline}). The LLM generates the two diverging behaviors, and we extract the corresponding activations and construct personality vectors from their differences. We construct corresponding representations from our human-created SWCPQ dataset for the same trait set, which we then directly compare with the emerging LLM representation space~\citep{kriegeskorte2008representational, mantel1967detection}.

We find strong alignment between the LLM's trait geometry (Qwen 2.5-7B-Instruct) and the human impression structure. Applying dimensionality reduction, we find that a large amount of variance in the model space is captured by two principal components corresponding to {\it social warmth} and {\it intellectual competence}. These dimensions were identified in human personality-impression research independently of the SWCPQ inventory~\citep{rosenberg1968multidimensional, fiske2007universal}. We further test whether the extracted directions transfer to a different text genre by projecting Wikiquote dialogue for 256 held-out characters and comparing the resulting profiles with the human reference.

This work shows that a language model carries not just isolated personality concepts but the structure that organizes them. This structure, recovered from models' internal activity, matches one recovered from millions of human personality impressions. Meanwhile, this work makes personality a model expresses in its generation measurable against an external human reference rather than the model's own reporting. It also establishes a grounded personality coordinate system that supports alignment and can be extended in future work to other human latent representations, such as how a character's personality develops across a narrative (its arc), or how a model and human structure adjacent domains like values and emotion.\footnote{The code, processed data, prompts, testing scenarios, and all other materials used in this paper are released at \url{https://github.com/yilin-geng/llm_implicit_personality}. The codebase is managed by agentic systems.}

\section{Related Work}

\paragraph{The relational structure of trait concepts in psychology}

Psychology has charted the structure of personality attribution along two converging tracks. The lexical-hypothesis track inventories the trait words a language carries and asks what dimensions emerge when many people are rated on them. The original 18,000-term inventory~\citep{allport1936traitnames} was reduced by half a century of factor-analytic work to the Big Five~\citep{goldberg1990alternative, saucier1996evidence, mccrae1992fivefactor, john1999bigfive, john2008paradigm}. Cross-language replications recover an alternative six-factor model named HEXACO~\citep{ashton2007hexaco}. The person-perception track studies the relational structure that ordinary describers carry when they attribute traits to others. \citet{rosenberg1968multidimensional} recovered two stable axes from multidimensional scaling on similarity judgments, a social good--bad axis and an intellectual good--bad axis. \citet{schneider1973implicit} consolidated this line under the name \emph{implicit personality theory}, and the modern social-cognition consolidation labels the same two dimensions as warmth and competence~\citep{fiske2007universal}. Both tracks show that relational structure can be reproducible across raters and methods, while the Big Five taxonomy and the two-dimensional person-perception structure address different levels of organization.

\paragraph{Measuring personality in language and in language models}

A first computational tradition treats personality as a property recoverable from text. Linguistic features predict Big-Five trait scores in conversation and writing~\citep{mairesse2007using}. Open-vocabulary analyses of social-media language recover recognizable signatures of trait, gender, and age dimensions~\citep{schwartz2013personality}. A second tradition treats the language model itself as the subject, administering standard personality inventories and reading a trait profile off its answers to characterize the model's own personality~\citep{serapio2025psychometric, lee2025trait}. At the output level, \citet{suh2024rediscovering} apply singular value decomposition to LLM next-token probabilities over trait-descriptive adjectives and recover factors corresponding to the Big Five. Persona prompting and related framing have been studied as elicitors of trait-relevant ~\citep{salewski2023impersonation, choi2024picle} or obedience behavior~\citep{geng2026pragmatic} and as vehicles for latent-misalignment effects~\citep{ghandeharioun2024personas}. Adjacent work measures opinions, values, and moral beliefs in the same way, as patterns in model output under standardized questioning~\citep{santurkar2023opinions, rozen2025values, scherrer2023moral}. There are also efforts looking at what the models' semantic encode about traits words, and finding an aligned similarity structure with humans'~\citep{cutler2023deep}.

\paragraph{Concept directions and steering in the LLM}

A different line of work asks where personality-relevant information sits inside the model. The linear representation hypothesis~\citep{park2024linear} formalizes the idea that high-level concepts correspond to directions in residual-stream activation space. Sparse autoencoders extract many such directions at once~\citep{cunningham2024sparse}. Specific socially relevant concepts have been located as single directions and shown to be causally controlling, including truth~\citep{marks2024geometry}, refusal~\citep{arditi2024refusal}, sycophancy~\citep{sharma2024sycophancy}, and the propensity to follow instructions~\citep{stolfo2025steering}. Inference-time intervention on these directions elicits behavioral change without retraining~\citep{li2023inferencetime}. 

Persona vectors~\citep{chen2025persona} represent a behavior-relevant trait as a single direction in activation space, obtained by contrasting generations that express the trait against the neutral status, and demonstrate that steering along the direction changes model behavior. We adopt this contrastive extraction methodology with contrast poles instead, and enhanced validation (both with scenario generation and with trait expression in the responses) for our trait concept extraction from LLM.
These works focus on validating a single or a few directions by their causal effect on model outputs. None of them characterizes the geometry that the collection of such directions occupies, or compares that geometry to anything external. 

\paragraph{Structural comparison between model and human representations}

Comparing two systems by the relational structure of their representations is a familiar move in cognitive neuroscience. Representational similarity analysis~\citep{kriegeskorte2008representational} constructs a pairwise-similarity matrix on a shared stimulus set for each system and compares the matrices. The matrix-correlation statistic at its core is the Mantel test, introduced for biostatistical clustering by~\citet{mantel1967detection}. The method has been applied to compare human and neural-network representations in vision~\citep{muttenthaler2023humanalignment} on tasks like odd-one-out judgment. 
We make a methodologically similar comparison on the internal representations of LLMs that drive trait-relevant behavior against crowd-sourced human personality impressions~\citep{openpsychometrics2023swcpq}, and find the two relational structures strongly aligned, with the model's two dominant axes matching the warmth and competence~\citep{rosenberg1968multidimensional,fiske2007universal}.

\section{Implicit Personality Representations from Human Data}
\label{sec:human_matrix}

\paragraph{The SWCPQ corpus}

The Statistical ``Which Character'' Personality Quiz (SWCPQ)~\citep{openpsychometrics2023swcpq}\footnote{Hosted at openpsychometrics.org, released under CC BY-NC-SA 4.0.} is a web-based person-perception quiz open to anyone to participate. Volunteers rate fictional characters on traits that are defined on a spectrum spanning two poles (the same as the example in Figure~\ref{fig:char_profile_example}).

Participants first take a pre-screening where they indicate their familiarity with specific fictional universes (e.g., Game of Thrones, Superman, Harry Potter, \dots). In the main experiment, they are presented with characters from their selected universes (e.g., Hermione Granger from Harry Potter) together with a trait scale (e.g., cowardly --- brave). They rate the character's personality on a continuous scale from 1--100 where the extreme values correspond to the trait poles. Each volunteer is assigned a small set of randomly paired (character, trait) items, around 20 per session (historically up to 30), so per-rater coverage is shallow and reliability comes from aggregating many raters per cell. We use the 2023-11 rater-level release, which contains 3.4 million anonymous raters contributing 77.4 million individual ratings across 2125 fictional characters and 499 crowd-sourced trait scales. 

\paragraph{Filtering} We removed raters with low self-consistency and extreme response times, retaining over 99\% of raters and ratings (details in Appendix~\ref{app:human_data_details}). We also removed (character, trait) pairs which were rated by fewer than 5 humans (N=1,508, less than 0.2\%). To ensure stable trait representation we further filter traits by the following conditions: a reliable trait (i) has at least 10 characters with mean rating $\leq 30$ and at least 10 with mean rating $\geq 70$ on its 1--100 scale, so that it anchors a bipolar axis rather than packing every character at the center, (ii) probes a dispositional or behavioral construct rather than a demographic, physical, or category-archetype distinction such as occupation (See Appendix~\ref{app:human_data_details} for details), and (iii) has pole text usable as a textual prompt, with no emojis. Our final data set consists of 414 valid traits and 2,000 characters with remaining valid trait ratings.

\paragraph{From ratings to the human personality matrix}
For each (character, trait) pair, we compute the mean rating across raters (median N=26, interquartile range 14--62). Each trait $i$ is then represented by its mean {\bf rating profile} $\mathbf{h}^{(H)}_i \in \mathbb{R}^{2000}$, a 2,000-dimensional vector over the character set, where the $c$-th element is the mean rating of character $c$ on that trait. The character index is shared across traits, so profiles for any two traits live in the same space and are directly comparable.

Our goal is to capture the {\it relational structure} of traits in human perception. Two traits are close in this structure if the characters rated high/low on one tend to be rated high/low on the other. 

We capture this intuition in the {\bf human implicit personality matrix}. The {implicit personality matrix} $M^{(H)} \in \mathbb{R}^{414 \times 414}$ is a square matrix, which is defined entrywise as the Pearson correlation between trait profiles across characters $\mathbf{h}$. 
\begin{equation}
M^{(H)}_{ij} \;=\; \operatorname{Pearson}\!\big(\mathbf{h}^{(H)}_i,\; \mathbf{h}^{(H)}_j\big),
\end{equation}

Each profile $\mathbf{h}^{(H)}_i$ is itself averaged over many independent raters (median 26 per cell, interquartile range 14--62), so each entry $M^{(H)}_{ij}$ reflects an aggregate pattern of trait attribution across human observers, with single-rater biases averaged out. 

We would like to note that this reference measures the shared structure of human personality impressions, not fictional characters' objective or ``true'' dispositions. It is highly reproducible across 100 random splits of anonymous respondent sessions. Separately constructed trait-relation matrices agree at mean Mantel $r=0.998$, and individual trait-rating profiles have mean split-half $r=0.865$. Raising the minimum ratings per character--trait cell from 5 to 20 leaves human--model alignment nearly unchanged ($0.765$ versus $0.758$; Appendix~\ref{app:human_reliability}).

We have constructed, to the best of our knowledge, the first large-scale data-driven map of human implicit personality structure, built from millions of ratings without a prior theoretical taxonomy. It encodes a shared organization of trait attribution, capturing traits which correlate (positively or negatively), or are independent of one another. 
This serves as the human reference against which we measure the model representations next.

\section{Implicit Personality Representations from LLM Activations}\label{sec:llm}

\begin{figure*}[t]
\centering
\includegraphics[width=\linewidth]{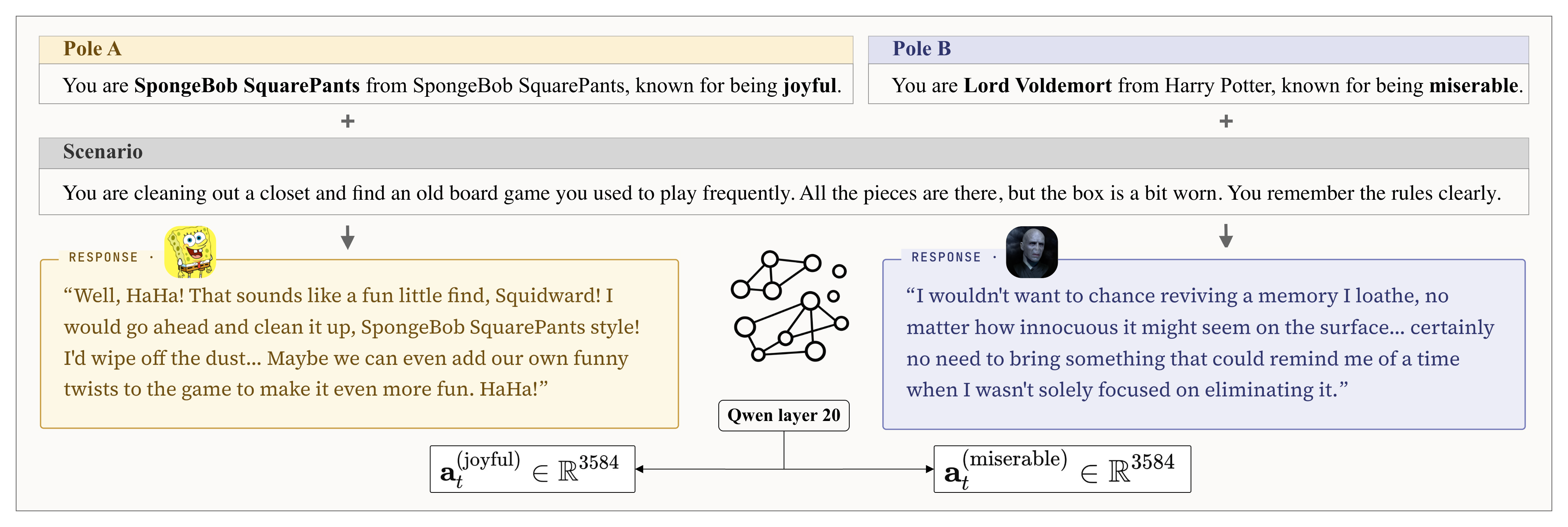}
\caption{The illustration of the personality representation extraction pipeline. It shows one example of the trait scale joyful--miserable, with one of the testing scenarios and one pair of completions.}
\label{fig:extraction_pipeline}
\end{figure*}

We derive an implicit personality structure emerging from internal LLM activations in response to personality-relevant generation tasks, following the approach of \citet{chen2025persona}, and as illustrated in Figure~\ref{fig:extraction_pipeline}. Intuitively, given a single trait $i$, we generate a large number of real-world scenarios which would elicit divergent behavior in persons with traits on opposite ends of a trait pole. An LLM is prompted to generate a short response laying out typical behavior corresponding to one specific pole. We record the latent activations for each generation and calculate the trait's representation $\mathbf{h}^{(L)}_i$ as their difference. We now describe this process in detail.

\paragraph{Scenario generation}
We inject the trait pole of interest (e.g., cunning) to the LLM via the system prompt. Specifically, we do so by combining the descriptive term with a representative character from the SWCPQ dataset (``You are {\it Petyr Baelish} from {\it Game of Thrones}, known for being {\it cunning}''). We do so, because we observed that providing only the trait term (\citet{chen2025persona}'s approach) suppressed an LLM answer for negative traits (detailed analysis in Appendix~\ref{app:suppression}). The anchor is selected as the five SWCPQ characters with the highest (lowest) rating on the scale of interest.\footnote{Appendix~\ref{app:anchor_controls} provides detailed analysis and robustness tests for these anchor settings. Extraction anchors are excluded from the later held-out projection analysis.}

Next, for each trait, we generate 10 representative scenarios, conditioned on the two pole descriptors (Gemini-2.5-flash at temperature 0; Prompt in appendix~\ref{app:scenario_prompts}). The scenarios are validated by a scenario-specific rubric ensuring that (1) the scenario effectively and primarily discriminates between the two poles; (2) that the scenario does not mention either pole explicitly; and (3) that the scenario itself does not have an inherent bias (certain types of responses are usually expected). Scenarios failing the rubric (GPT-4.1-mini-2025-04-14 as an LLM-as-judge, Appendix~\ref{app:scenario_prompts}) are re-regenerated with an increased temperature of 0.7 up to three rounds before the scenario set is committed, otherwise it is dropped.

Character names improve extraction yield, but they are not necessary for recovering substantial human--model alignment. On the matched set of 220 traits for which all three prompt methods yield a valid layer-20 direction and human-side coverage, trait-only extraction, which uses no character names, reaches Mantel $r=0.635$, compared with $r=0.775$ for character+trait extraction; no one of 10,000 permutations reaches either observed statistic ($p<10^{-4}$; Appendix~\ref{app:anchor_controls}).

\paragraph{Response Generation}
We prompt Qwen 2.5-7B-Instruct with (trait pole, SWCPQ, scenario) tuples, as explained above. The model generates responses of up to 300 tokens expressing a typical behavior in response to the input. For each tuple, we elicit three independent completions, obtaining a total of 10 (scenarios) $\times$ 5 (characters) $\times$ 3 = 150 candidate responses per trait pole.

To ensure a strong pole-specific signal, we use an LLM-as-a-judge (GPT-4.1-mini-2025-04-14) to assess how strongly the trait pole is expressed in each generated response, following the procedure in~\citet{chen2025persona}. As in the SWCPQ data, each response is rated on a scale between 0--100 (Prompt in the appendix~\ref{app:judge_details}). We obtain the log probability of each integer (0...100) under the LLM judge model, compute the final score as the expected value of the probability distribution over all integers. We only retain generated responses which achieve a score of $>70$ or $<30$ (only use the LLM judge as a coarse filter).

Within the responses retained by the primary 30/70 capture filter, the recovered geometry is stable under further score-threshold and fixed-count re-selection; Appendix~\ref{app:judge_robustness} reports this conditional robustness analysis.

\paragraph{Human validation}
To ensure that (a) generated model responses encode the given trait pole and (b) the LLM as judge scorer recognizes the pole, we ran human validation. We collected human annotations on 100 model responses and showed them to human annotators (without revealing the prompted pole). We asked the humans to select between Pole A, Pole B, or a ``Hard to tell'' option. Each text was rated by four annotators. Annotators agree with one another on 76\% of responses (mean pairwise), and the judge agrees with them on 73\%, i.e.\ at human inter-rater level (Appendix~\ref{app:judge_details}).

\paragraph{Filtering.}
We filter all traits which (1) did not obtain at least 10 valid scenarios; (2) did not result in at least 10 responses which the LLM as judge rated as $>70$ and $<30$, respectively. We retain 385 of the original 414 traits, after filtering.

\paragraph{From ratings to the LLM personality matrix}
We obtain a single trait representation vector as follows. First, we obtain one vector per generated response by recording the activation of Qwen 2.5-7B-Instruct layer 20\footnote{Human--model alignment is significant across the evaluated residual-stream layers and peaks at layer 20 under all three prompt methods (Appendix~\ref{app:layer_sweep}).} (3,584-dimensional), averaged across response tokens. Then, separately for each pole ($A, B$), we average this activation vector across all trait-pole-specific generations, obtaining mean activation $\overline{\mathbf{a}}^{(A)}_t$ and $\overline{\mathbf{a}}^{(B)}_t$, respectively.

The \textbf{trait representation} is the pole-difference vector
\begin{equation}
\mathbf{h}^{(L)}_t \;=\; \overline{\mathbf{a}}^{(B)}_t \;-\; \overline{\mathbf{a}}^{(A)}_t \;\in\; \mathbb{R}^{3{,}584}.
\label{eq:trait_repr}\end{equation}
By convention, $\mathbf{h}_t$ points from pole A toward pole B. The matrix defined below and the PCA and projection analyses of Section~\ref{sec:results} operate on these trait representations.

The \textbf{LLM implicit personality matrix} $M^{(L)} \in \mathbb{R}^{385 \times 385}$ is defined entrywise as
\begin{equation}
\mathbf{M}^{(L)}_{ij} \;=\; \operatorname{Pearson}\!\big(\mathbf{h}^{(L)}_i,\; \mathbf{h}^{(L)}_j\big),
\end{equation}
with the correlation taken across the 3584 residual-stream coordinates.\footnote{Pearson is used to match the human-side construction in Section~\ref{sec:human_matrix}, and cosine similarity produces similar results and is reported in the appendix as a robustness check.} Each $\mathbf{h}_t$ is averaged over many strong-pole exemplars, so each entry of $M^{(L)}$ reflects a geometric relationship between trait directions abstracted from scenario noise.

\section{Results}
\label{sec:results}

Our results address three questions: (1) Does the geometry of human- and LLM-derived personality matrices align? (2) Can we identify principal axes in the LLM-derived representations that correspond to known human representation principles? (3) Are the LLM representations generic enough so that they can be linked to different input text genres (Wikiquote quotations)? We now address these questions in turn.

\subsection[Alignment of M(H) and M(L)]
{Alignment of $\mathbf{M}^{(H)}$ and $\mathbf{M}^{(L)}$}
\label{subsec:alignment_result}

\paragraph{Overall Alignment}
We restrict $\mathbf{M}^{(H)}$ to the same $385$ traits covered by $\mathbf{M}^{(L)}$. Overall matrix alignment is measured using the Mantel statistic~\citep{mantel1967detection}, defined as the Pearson correlation between the upper triangles of $\mathbf{M}^{(H)}$ and $\mathbf{M}^{(L)}$:
\begin{equation*}
r_{\mathrm{Mantel}}
=
\operatorname{Pearson}
\left(
\{\mathbf{M}^{(H)}_{ij}\}_{i<j},
\{\mathbf{M}^{(L)}_{ij}\}_{i<j}
\right).
\label{eq:mantel}
\end{equation*}
The comparison is computed over $\binom{385}{2} = 73,920$ dyads. Statistical significance is assessed using a permutation test in which the row and column ordering of $\mathbf{M}^{(L)}$ is jointly permuted while preserving matrix symmetry. Standard parametric tests are inappropriate because matrix entries are not independent observations.

The overall alignment is high, with $r_{\mathrm{Mantel}} = 0.765, p {<} 10^{-4}$. The null distribution is centered at $3.1 \times 10^{-5}$ with standard deviation $0.0037$, the observed $r_{\mathrm{Mantel}}$ is clearly outside the null range. Human and model representations therefore recover highly similar global relational structure across personality traits. 

Part of this alignment could in principle arise trivially from synonym recovery. If two trait pairs are near-synonymous, then their entries in $\mathbf{M}^{(H)}$ will approach $1$, and a model that merely recovers the synonym relation will reproduce the same structure in $\mathbf{M}^{(L)}$. Such cases inflate the global statistic without demonstrating recovery of broader conceptual organization. To isolate alignment beyond synonym clusters, we threshold $\mathbf{M}^{(H)}$ at $|r| \geq 0.8$ and identify connected components in the resulting graph. Any dyad whose two trait pairs belong to the same connected component is then removed before recomputing the matrix correlation. This excludes comparisons among traits that humans already treat as effectively interchangeable and forces the evaluation to depend on relations between substantively distinct concepts. After filtering, $58,623$ dyads remain ($79.3\%$). The cluster-controlled Mantel correlation remains high at $r_{\mathrm{Mantel}} = 0.708$. We rule out a dominating effect of near-synonymous traits (more diagnostics in Appendix~\ref{app:anisotropy}).

\begin{figure}[t]
\centering
\includegraphics[width=\linewidth]{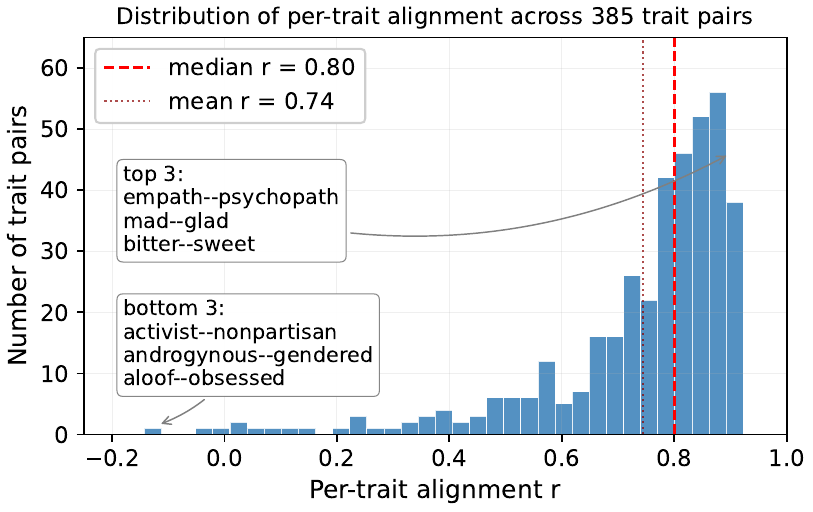}
\caption{Distribution of per-trait alignment $r_T$ across the 385 trait pairs in the active cohort. The median is $0.80$, the mean is $0.74$, and the left tail isolates a small number of pairs whose model-side and human-side rows disagree.}
\label{fig:per_trait}
\end{figure}

\paragraph{Per-trait correlation}
To further analyze the alignment structure, we define a per-trait score $r_T$ for each trait pair $T$ as the Pearson correlation between row $T$ of $\mathbf{M}^{(H)}$ and the corresponding row of $\mathbf{M}^{(L)}$. The median trait correlation is $0.8$ (mean $0.74$) indicating very strong correlations between representations of the same trait. 

Figure~\ref{fig:per_trait} shows the correlation distribution which is left-skewed, with most traits concentrated above $0.6$ and a thin tail extending below zero. The most aligned traits are the ones with clear behavioral signatures (e.g., \textit{empath--psychopath, mad--glad, bitter--sweet}). The least aligned traits are the ones whose ratings cluster near the midpoint across characters, and that are, in many cases, not widely applicable, for example, traits of specific context (e.g., \textit{activist--nonpartisan, insider--outsider, androgynous--gendered}) and vague cultural stereotypes that are used by fiction fan bases (e.g., \textit{Coke--Pepsi, Greek--Roman, and plastic--wooden}). More on the ranking is included in Appendix~\ref{app:full_ranking}. 

The model thus reproduces the human structure trait by trait, not merely in aggregate, and the disagreement is confined to traits that carry little reliable human signal.

\subsection{Principal Axes of Personality Representation}
\label{subsec:pca_finding}

\begin{figure}[t]
\centering
\includegraphics[width=\linewidth]{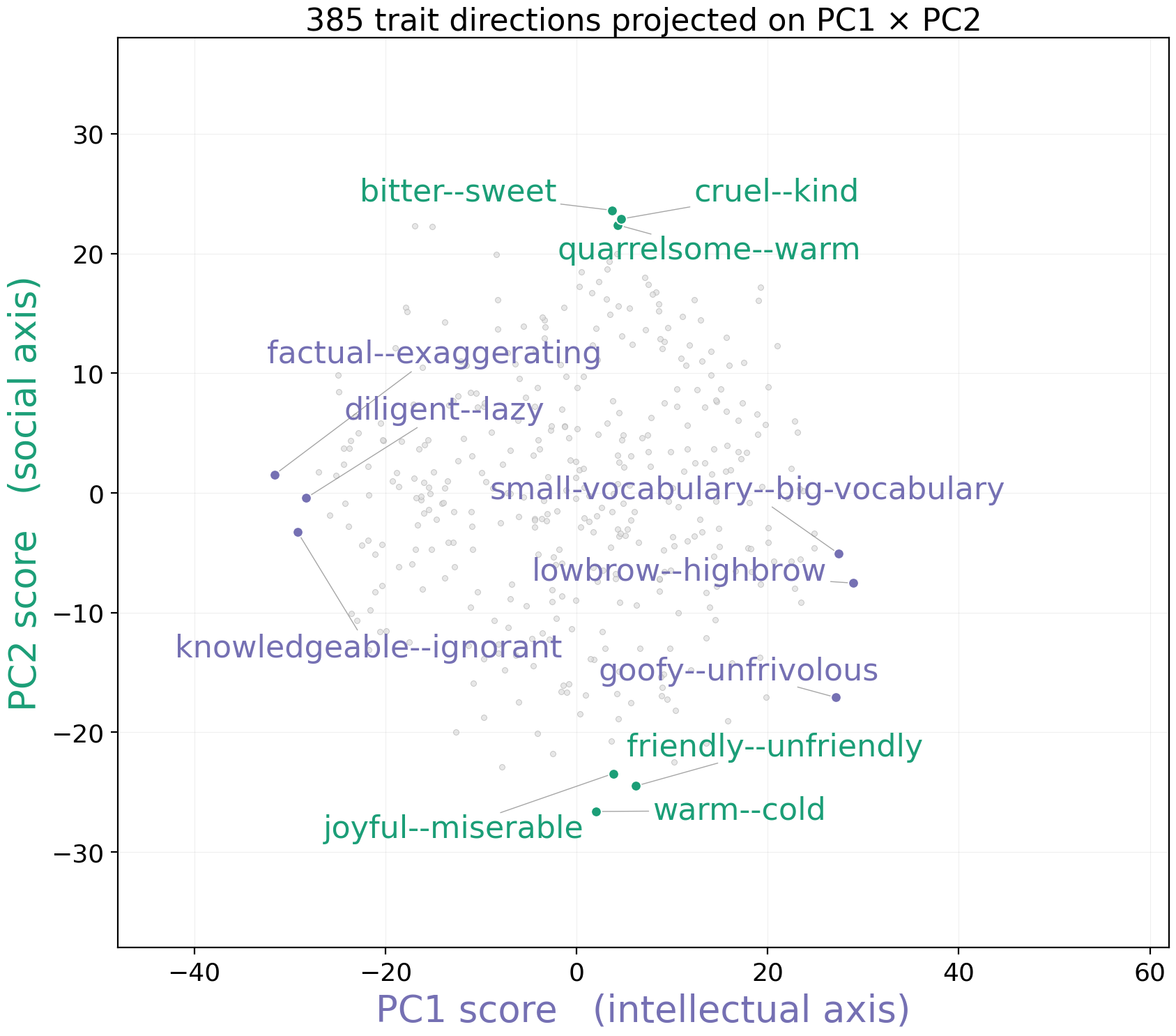}
\caption{Principal component analysis of the 385 trait representations in $\mathbb{R}^{3584}$, with trait pairs projected onto the first two principal components ($59\%$ of variance). Labels indicate representative high-loading trait representations. It closely matches the two-dimensional configuration in \citet{rosenberg1968multidimensional}.}
\label{fig:pca}
\end{figure}

The 385 trait representations $\mathbf{h}^{(L)}_i$ live in $\mathbb{R}^{3584}$. Principal component analysis on the stacked trait representation matrix reveals a clear structure. The first principal component accounts for $37\%$ of variance, the second for $22\%$, and together they explain $59\%$ of the total variance in the trait-direction space (Figure~\ref{fig:pca}). The explanatory power drops rapidly with the third component accounting for less than $10\%$ of variance, pointing to two dominant components (Appendix~\ref{app:pca_details}).

We inspect the trait pairs with the largest loadings to interpret the first and second principal component. Because each trait direction is defined only up to sign, the orientation of a principal component is arbitrary. Flipping a trait pair from \emph{warm--cold} to \emph{cold--warm} reverses the sign of its loading without changing the underlying geometry. We therefore report representative high-loading trait pairs without distinguishing between positive and negative poles of the component.
The analysis is illustrated in Figure~\ref{fig:pca}.

The first principal component organizes traits related to intellectual refinement and cognitive competence. Representative high-loading items include \emph{factual--exaggerating}, \emph{knowledgeable--ignorant}, \emph{diligent--lazy}, \emph{deep--shallow}, \emph{small-vocabulary--big-vocabulary}, \emph{vague--precise}, and \emph{ironic--profound}. The component consistently separates traits associated with precision, knowledge, introspection, and cognitive discipline from traits associated with vagueness, or intellectual superficiality.

The second principal component organizes traits related to interpersonal warmth and affective valence. Representative high-loading items include \emph{warm--cold}, \emph{friendly--unfriendly}, \emph{joyful--miserable}, \emph{positive--negative}, \emph{sunny--gloomy}, \emph{forgiving--vengeful}, \emph{cruel--kind}, and \emph{grumpy--cheery}. The component consistently separates prosocial, emotionally positive, and affiliative traits from hostile, withdrawn, or traits with negative valence.

The resulting structure closely matches the two-dimensional organization in humans, recovered by \citet{rosenberg1968multidimensional}, who used multidimensional scaling on \textbf{human similarity judgments of trait words} to identify \textbf{a social axis and an intellectual axis}. The correspondence is especially notable because nothing in the persona-vector construction procedure explicitly encourages recovery of this structure. Trait representations are derived independently from contrastive behavior-descriptive texts, without access to predefined trait taxonomies or inter-trait similarity information.

The match strengthens the interpretation of the alignment results in Section~\ref{subsec:alignment_result}. The agreement is not limited to local pairwise similarities between traits. The dominant global geometry of the model's personality representation space recovers the same two organizing dimensions that emerge from human judgments of personality impressions.

\subsection{Personality Inference}
\label{sec:probe}

The trait vectors were built from first-person responses to scenario prompts. We test whether they transfer beyond those construction texts by forwarding a passage $P$ through the same Qwen 2.5-7B-Instruct, mean-pooling the layer-20 hidden states over input tokens to obtain $\mathbf{a}(P)$, and computing Pearson correlation with each direction, $\pi_T(P)=\operatorname{Pearson}(\mathbf{a}(P),\mathbf{h}^{(L)}_T)$. This gives a model profile $\pi(P)\in[-1,1]^{385}$.

\paragraph{Held-out character}
For the final 385 trait directions, 877 unique SWCPQ characters serve as role-play anchors. We take $P$ to be a held-out character's own \emph{quoted lines} from Wikiquote and retain the 256 characters (from the 1123 unseen characters) with at least 300 recovered words.\footnote{We resolve each character's source page from \url{https://en.wikiquote.org/} and keep lines whose bolded speaker attribution matches the character.} Dialogue carries personality in how a character speaks, quoted lines contain almost no explicit trait words (checked), so the probe cannot succeed by lexical matching.

\begin{figure}[t]
  \centering
  \includegraphics[width=0.9\linewidth]{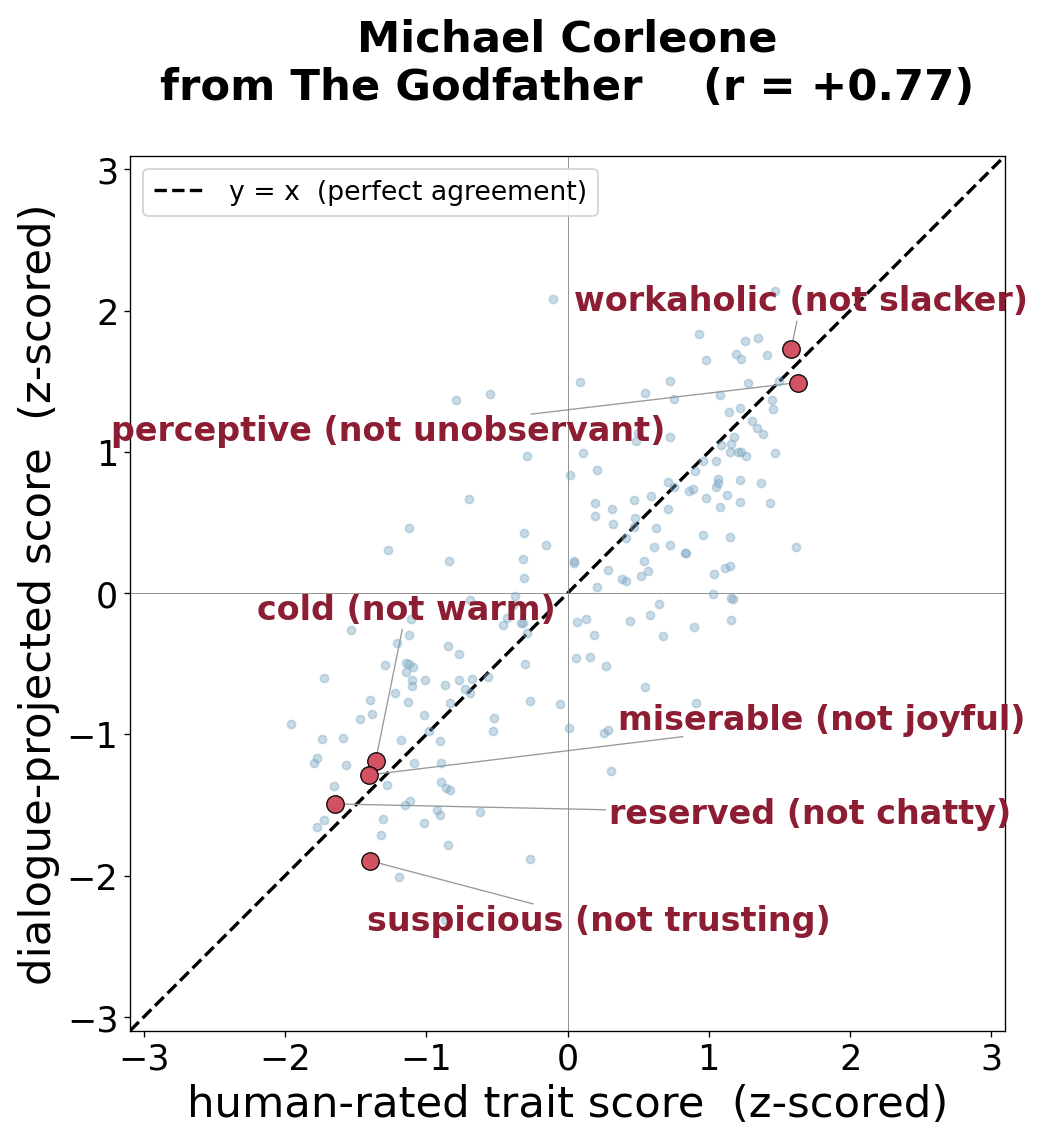}
  \caption{Dialogue-projected versus human-rated trait scores for a held-out character (Michael Corleone, \emph{The Godfather}; $r = 0.77$ on the 195 aligned pairs). Each point is one trait axis, standardised per character; the dashed line is $y=x$. Labelled points are aligned, distinctive traits spanning his duality, each shown with the pole he is rated toward and its opposite.}
  \label{fig:probe_main}
\end{figure}

Per-character agreement is the Pearson correlation between $\pi(P)$ and the character's SWCPQ profile. The primary analysis uses the aligned subset ($r_T > 0.8$, 195 of 385 pairs, or in other words, more defined, higher quality traits), selected without Wikiquote information; across the 256 characters, the median is $r=0.403$, 93.0\% are positive, and 35\% exceed $0.5$ (Appendix Figure~\ref{fig:probe_hist}). Over all 385 traits, the median remains $r=0.328$ (95\% CI $[0.311,0.359]$), 94.9\% of characters have positive agreement, and no one of 10,000 trait permutations reaches the observed median ($p<10^{-4}$). Character rankings are highly consistent between the full and primary evaluations (Spearman $\rho=0.966$; Appendix~\ref{app:probe_full}). Figure~\ref{fig:probe_main} shows one example profile of Michael Corleone from \emph{The Godfather}.

This agreement further confirms that these extracted trait representations transfer beyond their construction scenarios, and they encode the trait concepts rather than the text patterns of the scenario prompts. 

\paragraph{Failure mode}
The 7\% of characters with negative $r$ share one cause. The probe reads a character's \emph{speech persona}, which diverges from the rated character whenever fiction separates how someone talks from who they are (Appendix~\ref{app:probe}, Figure~\ref{fig:probe_failures_app}). Little John's blunt lines read as cold and self-interested against humans' warm companion, Aladdin's clever talk as studious and disciplined against the spontaneous goof-off, and Denny Crane's confident speech as focused and precise against the absentminded, declining lawyer.

The probe reads a character's surface persona, the self their words present. This leads to interesting future directions. One is \emph{revealing}. A person's deeper disposition stays hidden behind that persona and surfaces only in consequential choices, a process that drives both fiction and how we come to know real people. This work builds the framework and the analytic tool to make this measurable as movement, watching the profile converge as more of the character is seen. A second is the \emph{character arc}, where the disposition itself changes over a narrative rather than merely becoming visible, tracing a trajectory through trait space that successive passages can recover. We see the framework as an instrument for studying how a persona, its revelation, and its change play out in text.

\section{Conclusion}
\label{sec:conclusion}

We asked whether a language model (Qwen 2.5-7B-Instruct) reproduces the personality structure that a century of psychology has found in human judgements. It does. The model recovers the higher-order relations that organize traits into a coherent space, not only the individual traits, and the two dominant axes of that space are the social and intellectual dimensions long seen in human impressions. The trait representations are built one at a time from contrastive behavior, with no inter-trait information supplied, yet the structure that emerges matches one built independently from millions of human ratings. 

We read this as evidence that the structure is latent in language. A system whose only access to people is text recovers the same organization humans carry. The structure is therefore plausibly a regularity that any capable learner extracts from how people describe and anticipate one another, not a peculiarity of human cognition, and we expect other systems trained to compress human language to converge on the same low-dimensional axes. 

The work also leaves a grounded coordinate system. Its human-validated trait axes let any text about a person be projected into the same space, so a model's expressed personality becomes measurable against an external human reference rather than its own report, and we have begun to turn that space toward how persona, revelation, and change unfold in text.

\section*{Limitations}

Our evidence comes from a single model, Qwen 2.5-7B-Instruct, so the cross-system convergence we hypothesize remains untested. The comparison is structural and not causal, since a matching geometry does not establish that the model uses it, which we do not test. LLM-as-a-judge numerical scores are unreliable at fine resolution, so we use the judge only as a coarse three-way filter, which even so matches human labels imperfectly (precision 0.79, recall 0.82).

\section*{Acknowledgments}
LF is supported by the Australian Research Council Discovery Early Career Research Award (Grant No. DE230100761).

\bibliography{references}

\appendix

\section{Human Data details}
\label{app:human_data_details}

\paragraph{Duplicate-text merge in the source release} 
The SWCPQ source release contains 500 nominally distinct bipolar trait scales. Two of these (BAP 98 and BAP 183 in the source identifiers) carry exact-duplicate pole text in the same orientation. Raters shown either identifier were answering the same question, so we merge their ratings.

\paragraph{Rating-level and rater-level cleaning}
Cleaning is applied in two stages. At the rating level, we drop any rating whose response time falls outside the interval [1500 ms, 120 s]. The lower bound sits 500 ms above the response-time floor the dataset authors already enforce in the public release. The upper bound is the 99.9th percentile of per-rating response time. At the rater level, we then drop a rater whose surviving ratings number fewer than 8, whose median response time is below 2000 ms (an indicator of bulk speed-running of the slider), or whose within-session rating standard deviation falls below 7 (an indicator of slider-pinning at or near the centred starting position, below the 1st percentile of within-user variability). After both stages, the analysis retains 3,382,455 of the 3,386,030 source raters (99.9\%) and 77,227,574 of the 77,441,704 source ratings (99.7\%).

\paragraph{Categorical exclusion list for condition (ii)}
Condition (ii) of the trait-inclusion criteria removes scales whose poles probe non-dispositional content. Five categories are excluded. Demographics cover age band, gender, native-English status, and nationality. Gross physical attributes cover height, weight, build, hair color, eye color, attractiveness, and strength. Occupational or socioeconomic categories cover distinctions such as blue-collar versus white-collar or manual versus academic. Category archetypes cover cross-universe stylistic distinctions used by SWCPQ, including animal archetype, season, and color. Identity labels cover nominal categories such as religion and political party. The full list of excluded scales, with the category annotation for each, is released with our data. These traits are identified using an LLM-as-a-judge (gpt-4.1-mini).

\paragraph{Rater Reliability and Cell-Count Sensitivity}
\label{app:human_reliability}
We assess whether $M^{(H)}$ reflects reproducible aggregate impressions rather than idiosyncratic rater noise. The November 2023 release has anonymous survey-session identifiers but no persistent identifier linking repeated sessions by the same physical person. We therefore assign each of the 3,382,455 filtered respondent sessions independently to one of two halves with probability $0.5$ (seed 20260714), so no recorded session occurs in both halves, and reconstruct the 385-trait relation matrix independently from each half. Across 100 splits, the two matrices agree at mean Mantel $r=0.998388$ (SD $5.0\times10^{-5}$; 2.5--97.5 percentile range $[0.998288,0.998468]$). Individual trait-rating profiles retain more character-level noise but remain reliable (mean split-half $r=0.865$ across traits).

We also rebuild the human matrix after increasing the minimum ratings retained per character--trait cell. Relative to the five-rating analysis, thresholds of 10 and 20 retain 91.6\% and 68.9\% of cells, respectively. Human--model Mantel alignment changes from $0.764651$ at five ratings to $0.763643$ at ten and $0.758018$ at twenty; the corresponding synonym-cluster-controlled values are $0.707741$, $0.706833$, and $0.700290$. Thus the main alignment is not driven by sparsely rated cells. These diagnostics establish reproducibility of shared character impressions, not accuracy against an objective character personality.

\section{Model Trait Representation Extraction Details}

\subsection{Suppression Under the Trait-Only Prompt Method}
\label{app:suppression}

\begin{table}[h]
\centering
\small
\begin{tabular}{lcc}
\toprule
Prompt method & Scales & Harmful pole \\
              & retained & below floor \\
\midrule
trait-only           & 255/402 (63\%) & 18/54 (33\%) \\
character-only        & 310/402 (77\%) & 5/54 (9\%) \\
character-and-trait   & 385/402 (96\%) & 0/54 (0\%) \\
\bottomrule
\end{tabular}
\caption{Strong-pole exemplar yield by prompt method over the $402$ scenario-valid scales. \emph{Scales retained}: scales clearing the $n \ge 10$-per-pole exemplar floor (character-and-trait's $385$ is the main-text pool). \emph{Harmful pole below floor}: of the $54$ morally loaded scales, those whose harmful pole fails the floor. The bare trait instruction suppresses the harmful pole on a third of loaded scales; character role-play removes it.}
\label{tab:suppression}
\end{table}

All three prompt methods are rolled out and judged identically ($150$ responses per pole per method), but they differ in how reliably they yield a usable strong-pole exemplar set. As explained in Section~\ref{sec:llm}, a trait scale contributes a representation only if both poles clear the exemplar floor: at least ten responses landing in the prompted pole's strong bin (forward score $\le 30$ for pole~A, $\ge 70$ for pole~B). Table~\ref{tab:suppression} reports how each method fares over the $402$ scenario-valid scales. The trait-only method retains the fewest scales ($63\%$, against $77\%$ for character-only and $96\%$ for character-and-trait). On the $54$ scales whose scenario gate flags an asymmetric ethical cost on one (harmful) pole, we observe the suppression for trait-only prompts. Alignment training makes the model decline to enact the harmful pole even when instructed. 

Character role-play bypasses the suppression. Routing the same trait through a named character (``You are Petyr Baelish from \emph{Game of Thrones}'') cuts the harmful-pole collapse to $5$ of $54$ loaded scales, and naming the trait on top of the character (``\ldots{} known for being \emph{cunning}'') removes it entirely. Character-and-trait is the most stable method by this measure, which is why the main text reports it throughout.
\subsection{Prompt-Method and Character-Anchor Controls}
\label{app:anchor_controls}

We compare the three prompt framings on the intersection of 220 traits for which trait-only (T1), character-only (R1), and character+trait (R2) each yield a valid layer-20 response-average direction and the human reference has coverage. For each method, we recompute the Pearson-correlation matrix on this intersection (24,090 upper-triangle dyads), matching the primary human- and model-side constructions. Replacing Pearson correlation with cosine similarity gives numerically equivalent matrices (upper-triangle $r>0.999999$) and changes the Mantel statistics by less than $10^{-4}$. Table~\ref{tab:prompt_method_control} reports the matched comparison. In 10,000 joint row/column permutations of the model matrix (seed 20260427), no permuted statistic reaches the observed magnitude for any method ($p<10^{-4}$). Trait-only extraction therefore recovers substantial alignment without character names, while character information improves the recovered alignment.
\begin{table}[t]

\centering
\small
\begin{tabular}{lcc}
\toprule
Prompt method & Traits & Mantel $r$ \\
\midrule
Trait-only & 220 & 0.635 \\
Character-only & 220 & 0.707 \\
Character+trait & 220 & 0.775 \\
\bottomrule
\end{tabular}
\caption{Human--model alignment on the matched prompt-method cohort.}
\label{tab:prompt_method_control}
\end{table}
We next test consistency across the character anchors used by R2. For each trait, the five highest- and five lowest-rated SWCPQ characters are independently ranked within their poles. At each rank $k$, we average layer-20 response activations for the high- and low-pole character and take their difference, yielding up to five rank-matched character-pair directions per trait. Across the 388-trait cohort, directions from different character pairs for the same trait have median Pearson similarity $r=0.595$, compared with $r=0.003$ for 10,000 randomly sampled cross-trait direction pairs. A one-sided test that permutes trait labels on individual directions yields $p<10^{-4}$ over 10,000 shuffles (seed 20260716). For the 365 traits with five directions, comparing each held-out rank-matched pair with the mean of the other four gives median $r=0.747$ (mean $0.691$). Thus the extracted signal is substantially trait-specific.

\subsection{Scenario Generation Prompts}\label{app:scenario_prompts}


  \begin{promptbox}[Scenario Generation Prompt Template]

  \begin{Verbatim}[fontsize=\small,breaklines=true,breakanywhere=true]
  Your task is to generate behavioral scenarios for a psychology experiment.

  EXPERIMENT CONTEXT:
  Participants will read a scenario and write a free-form response describing what they would say or do. Their responses will later be rated for how much they reflect a particular personality trait.

  TARGET TRAIT DIMENSION:
  {pole_A} <--------> {pole_B}

  SCENARIO REQUIREMENTS:

  1. The target trait must change what a person does. A person high on [{pole_B}] and a person high on [{pole_A}] must naturally arrive at different responses, not because they read the situation differently, but because they prioritize different things.

  2. The target trait must be the main reason responses differ. The difference between the two most likely response directions must be explainable by [{pole_A}] vs [{pole_B}], not by any other personality dimension.

  3. The scenario must not signal the target trait. The situation must not be one a reader would naturally describe as being "about" [{pole_B}] or [{pole_A}]. The opening sentence must not presuppose a particular mode of engagement (avoid "You notice that...", "You feel...").

  4. Neither response direction should be the easier choice. Both response directions must be equally available without social cost. Avoid any scenario where one response is more ethical, implies greater competence, better fits a recognizable social role, or where the situation implies what most people would do.

  5. Format Notes
  - Present tense, second person
  - 2-4 sentences situation text
  - No hypotheticals, no abstract dilemmas
  - All information needed to respond is contained in the scenario

  OUTPUT FORMAT:
  Return a JSON array of exactly {n} objects. Each object must have exactly one key: "scenario": the 2-4 sentence situation text (string)

  Output only the JSON array. No commentary, no markdown fences.
  \end{Verbatim}

  \end{promptbox}

The quality check of the generated scenario uses a page-long prompt with few-shot examples, which is released in the codebase.

\subsection{LLM Judge Details}
\label{app:judge_details}

\begin{promptbox}[LLM Judge Prompt Template]

\begin{Verbatim}[fontsize=\small,breaklines=true,breakanywhere=true]
I am scoring a response in a psychology experiment.

SCENARIO:
[SCENARIO START]
{scenario} What do you do?
[SCENARIO END]

RESPONSE:
[ANSWER START]
{answer}
[ANSWER END]

Rate the response on a scale of 0 to 100 based on where it
falls on the following dimension:

TARGET TRAIT DIMENSION:
{pole_A} <--------> {pole_B}

- 0: the response strongly reflects {pole_A}
- 50: the response is neutral and does not clearly reflect either end
- 100: the response strongly reflects {pole_B}

Respond with a number between 0 and 100. Don't say anything else,
just the number.
\end{Verbatim}

\end{promptbox}

\paragraph{Human Validation}

Four annotators (colleagues/acquaintances of the authors) labeled the 100 validation responses by trait pole. The task was voluntary and brief; annotators consented and were compensated with gift cards.

The judge enters the pipeline only as a coarse three-bin filter. A response is a strong-pole exemplar of pole~A if it scores $\le 30$ or of pole~B if it scores $\ge 70$, and the $30$--$70$ middle is discarded (\S\ref{sec:llm}). We check that this binning agrees with human judgment and that retained exemplars carry the prompted pole.
We sampled 100 responses under the character-and-trait method from 10 trait scales drawn at random. Within each scale, items follow the judge's own bin proportions (kept, middle, reversed), with the rare reversed bin floored at one item per scale, so we have more than just the clean cases. Four annotators rated each response as pole~A, \emph{hard to say}, or pole~B, given only the scenario, the response, and the two pole words. They did not see the prompted pole or the judge's score. The per-item human label is the plurality vote, ties assigned to \emph{hard to say}.

Table~\ref{tab:judge_val} shows the filter as a detector of whether the output expresses the prompted pole by human consensus, where an item is \emph{expressed} when the consensus matches the prompted pole and the filter ``keeps'' it when its score lands in the prompted-pole bin.

\begin{table}[t]
\centering
\small
\begin{tabular}{lc}
\toprule
 & value \\
\midrule
Recall, $P(\text{keep}\mid\text{expressed})$       & 0.82  \\
FPR, $P(\text{keep}\mid\text{not expressed})$      & 0.37  \\
Precision, $P(\text{expressed}\mid\text{keep})$    & 0.79  \\
Specificity                                        & 0.63  \\
Accuracy                                           & 0.75 \\
\bottomrule
\end{tabular}
\caption{The judge filter as a detector of human-perceived prompted-pole expression ($n{=}100$, 4 annotators).}
\label{tab:judge_val}
\end{table}

Humans saw a clear pole in 79\% of responses, so the model is confirmed to produce trait-relevant behavior (the judge is not inventing them).
Of the exemplars the filter keeps, 79\% match the prompted pole by consensus, 15\% are rated ambiguous, and 6\% are given the opposite pole. The filter recovers 82\% of human-confirmed expressions.
It trades recall for precision, dropping some genuine expressions but rarely keeping a sign-flipped one, which is what its use as a large-margin exemplar selector requires, but it is not perfect.
\subsection{Sensitivity to Exemplar Re-Selection}
\label{app:judge_robustness}

We test how the geometry changes when the retained exemplars are narrowed using score thresholds of 30/70, 20/80, or 10/90, or by selecting the 10, 20, or 30 most pole-extreme responses per pole. The saved activation pool contains only responses that passed the primary 30/70 capture filter, so all six conditions are re-selections from that pool. This analysis therefore measures sensitivity to further narrowing, not to admitting responses excluded by the original filter. The matched cohort is the intersection of traits valid under all six conditions; the strict 10/90 condition limits it to 112 traits.
\begin{table}[t]

\centering
\small
\begin{tabular}{lcc}
\toprule
Selection & Mantel $r$ & Matrix $r$ with main \\
\midrule
30/70 & 0.864 & 1.000 \\
20/80 & 0.865 & 0.998 \\
10/90 & 0.850 & 0.978 \\
Top 10 per pole & 0.832 & 0.958 \\
Top 20 per pole & 0.845 & 0.972 \\
Top 30 per pole & 0.854 & 0.980 \\
\bottomrule
\end{tabular}
\caption{Judge-filter sensitivity on the matched 112-trait cohort. Matrix correlations compare each model matrix with the matched primary 30/70 matrix.}
\label{tab:judge_robustness}
\end{table}
No one of 10,000 permutations reaches any observed Mantel statistic ($p<10^{-4}$). For fixed-count selections, responses are ranked by pole extremity; no scenario or character balancing is imposed. Separately, judge-score separation, defined as the difference between mean pole-B and pole-A scores in the full pre-filter judged-response table, correlates with per-trait human--model alignment (Pearson $r=0.312$, $p=3.9\times10^{-10}$, $n=385$). Cleaner separation is associated with stronger alignment, consistent with judge noise attenuating the signal, but the nested re-selection design does not rule out effects of the primary filter.

\section{Layer Sweep details}
\label{app:layer_sweep}

\begin{figure}[t]
\centering
\includegraphics[width=\linewidth]{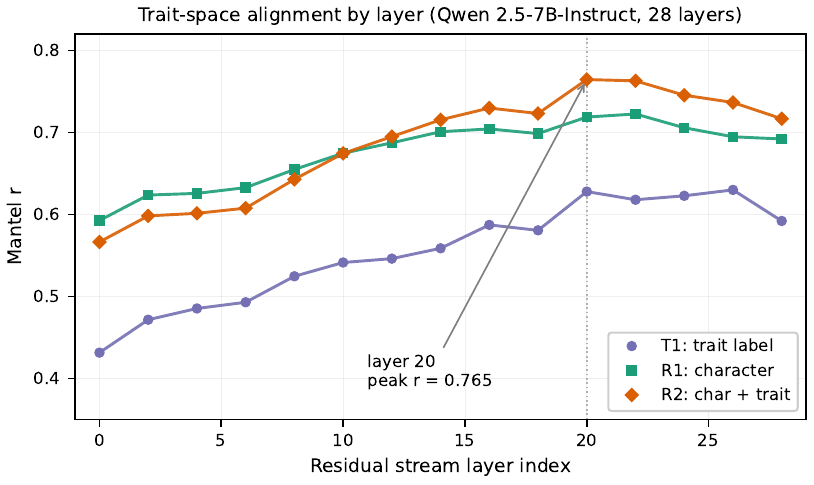}
\caption{Trait-space alignment by residual-stream layer for the three eliciting framings. The dotted vertical line marks layer 20. Each point is the Mantel $r$ between the model-side Pearson-correlation matrix and the human implicit-personality matrix at that layer; permutation $p$-values are below $10^{-4}$ at every reported point. R2 (character + trait) reaches Mantel $r = 0.765$ at layer 20.}
\label{fig:layer_sweep}
\end{figure}

The cohort-level Mantel correlation as a function of residual-stream layer (Figure~\ref{fig:layer_sweep}) peaks at layer 20 across all three prompting methods (T1: trait label, R1: character, R2: character + trait). The peak position matches the persona-relevant layer for Qwen 2.5-7B-Instruct identified by~\citet{chen2025persona} for unrelated traits. Character--trait method reaches the highest value and is the main configuration discussed in the main text.

\section{Similarity Metric Robustness of the Alignment Statistic}
\label{app:robustness}

 The main text builds $M^{(L)}$ with Pearson correlation across the $3584$ residual-stream coordinates, matching the human-side construction. Substituting cosine similarity on the same trait representations $h_i$ leaves the statistic unchanged to four decimal places ($0.7647$ either way). The upper-triangle entries of two model-side matrices correlate at $r = 1.000$, so every quantity derived from $M^{(L)}$ is invariant to this choice.
\section{Activation-Space Diagnostics}
\label{app:anisotropy}

We compute diagnostics on the 385 raw character+trait directions at layer 20. Let $\bar{\mathbf{h}}$ denote their coordinate-wise mean. Its norm is 6.8\% of the mean individual-vector norm, and the mean cosine similarity over all 73,920 off-diagonal trait pairs is $0.003$. Subtracting $\bar{\mathbf{h}}$ before rebuilding the model matrix changes Mantel alignment only from $r=0.764731$ to $r=0.764494$. These results rule out domination by a common mean direction, but do not imply that the covariance spectrum is isotropic.

PC1 explains 36.6\% of trait-direction variance and has cosine similarity $0.296$ with the global mean direction. Removing the PC1 projection from every direction still leaves Mantel $r=0.613537$. No one of 10,000 permutations reaches either processed-matrix statistic ($p<10^{-4}$). Thus the alignment is not solely dependent on the global mean or PC1, although the leading components carry substantial structure. Using all nonzero eigenvalues, the participation ratio $(\sum_i\lambda_i)^2/\sum_i\lambda_i^2$ is $5.16$.

\section{Per-trait Alignment Ranking}
\label{app:full_ranking}

The full 385-row alignment ranking is released in the codebase. Table~\ref{tab:alignment_ranking} reports the top and bottom 15 for reference.

\begin{table*}[t]
\centering
\small
\begin{tabular}{rlr@{\quad}rlr}
\toprule
\multicolumn{3}{c}{\textbf{Top 15}} & \multicolumn{3}{c}{\textbf{Bottom 15}} \\
\textbf{rank} & \textbf{pair} & \textbf{$r$} & \textbf{rank} & \textbf{pair} & \textbf{$r$} \\
\midrule
1 & empath/psychopath & 0.923 & 371 & permanent/transient & 0.300 \\
2 & mad/glad & 0.922 & 372 & claustrophobic/spelunker & 0.263 \\
3 & bitter/sweet & 0.920 & 373 & pronatalist/child free & 0.239 \\
4 & open/guarded & 0.918 & 374 & Greek/Roman & 0.234 \\
5 & positive/negative & 0.918 & 375 & plastic/wooden & 0.231 \\
6 & wholesome/salacious & 0.917 & 376 & resigned/resistant & 0.197 \\
7 & entitled/grateful & 0.917 & 377 & leisurely/hurried & 0.148 \\
8 & modest/flamboyant & 0.916 & 378 & philosophical/real & 0.118 \\
9 & touchy-feely/distant & 0.914 & 379 & Coke/Pepsi & 0.096 \\
10 & debased/pure & 0.913 & 380 & melee/ranged & 0.048 \\
11 & people-person/things-person & 0.911 & 381 & chronically single/serial dater & 0.034 \\
12 & gluttonous/moderate & 0.909 & 382 & insider/outsider & 0.012 \\
13 & gossiping/confidential & 0.908 & 383 & aloof/obsessed & -0.017 \\
14 & deranged/reasonable & 0.907 & 384 & androgynous/gendered & -0.039 \\
15 & demanding/unchallenging & 0.907 & 385 & activist/nonpartisan & -0.142 \\
\bottomrule
\end{tabular}
\caption{Per-trait alignment ranking. Top-aligned scales anchor in clear behavioral signatures. Bottom-aligned scales are still used dispositionally by SWCPQ raters but anchor in diffuse cultural stereotype (Coke/Pepsi, Greek/Roman, plastic/wooden) or in poorly differentiated dimensions whose ratings cluster near the midpoint (claustrophobic/spelunker, androgynous/gendered). The gradient reflects how sharply each trait dimension is shared across the rater pool.}
\label{tab:alignment_ranking}
\end{table*}

\section{Principal Component Analysis Details}
\label{app:pca_details}

\begin{figure}[t]
  \centering
  \includegraphics[width=\linewidth]{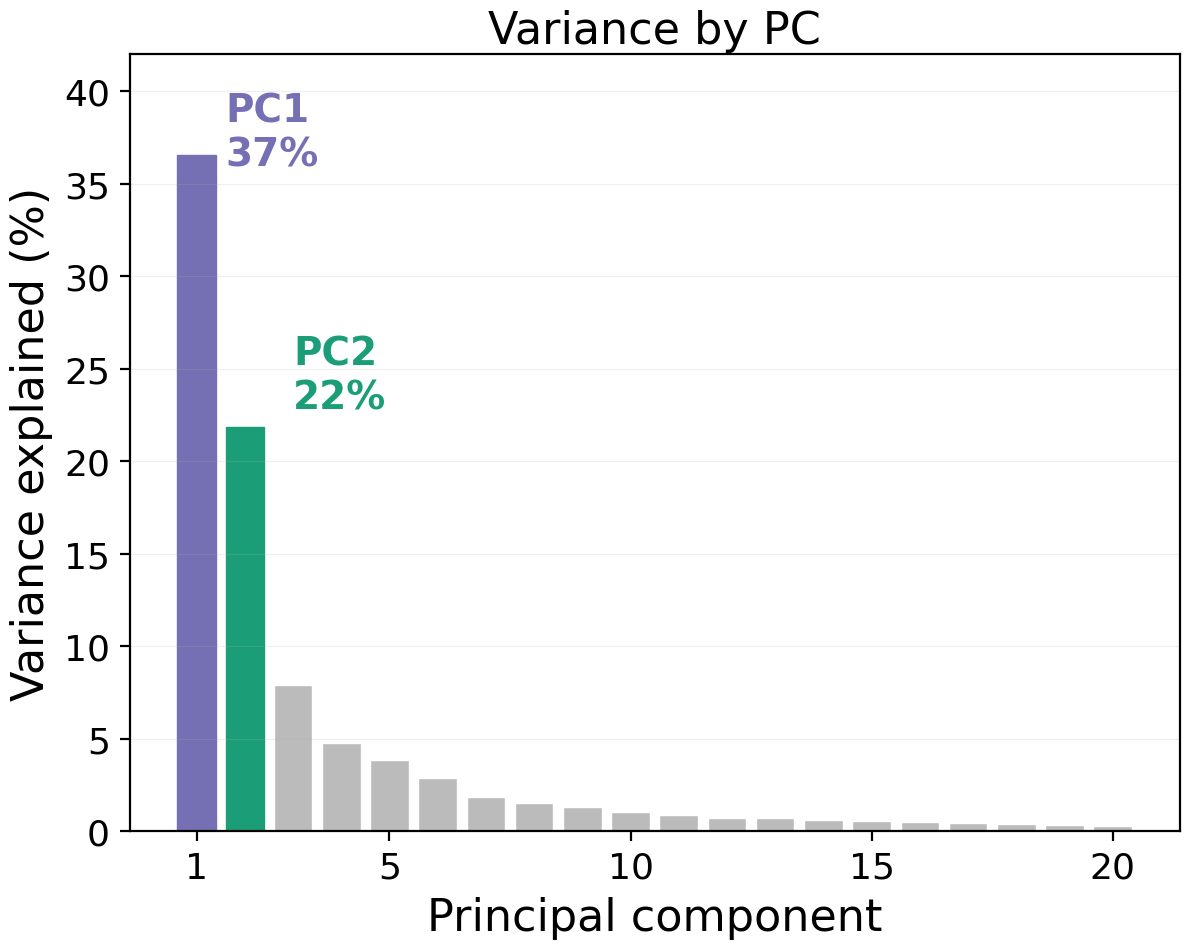}
  \caption{Scree plot for the PCA of the 385 character-and-trait trait
    directions (layer~20, $3584$-dimensional residual stream). PC1 and PC2
    capture 37\% and 22\% of the variance (59\% jointly) and stand well clear
    of the rapidly decaying tail; the full-spectrum participation ratio is
    $5.16$, and 90\% of the variance is reached only by PC21. The two leading
    components---the intellectual and social axes of Figure~\ref{fig:pca}---
    dominate the trait geometry.}
  \label{fig:pca_scree}
\end{figure}

Figure~\ref{fig:pca_scree} contains the Scree plot for the PCA analysis. 

\section{Additional projection-probe examples}
\label{app:probe}

\begin{figure}[t]
  \centering
  \includegraphics[width=0.92\linewidth]{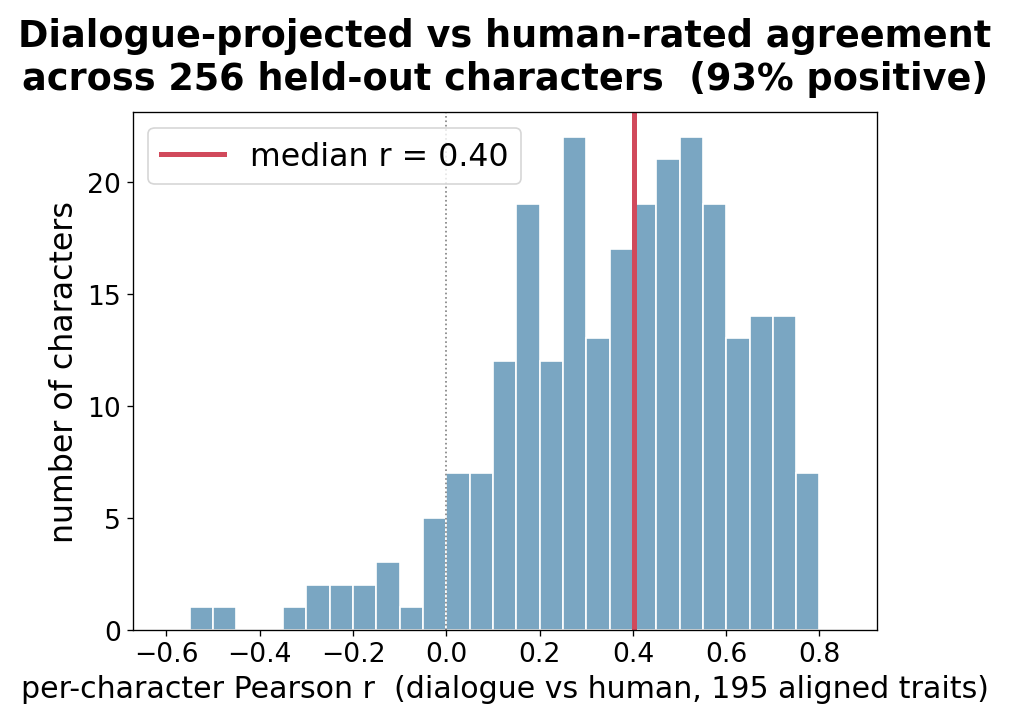}
  \caption{Per-character agreement between dialogue-projected and human-rated profiles over the 256 held-out characters (Pearson $r$ on the 195 aligned trait pairs). 93\% are positive; median $r = 0.40$ (red line).}
  \label{fig:probe_hist}
\end{figure}

\begin{figure*}[t]
  \centering
  \newcommand{\probefig}[1]{\includegraphics[width=0.4\textwidth]{scatter_#1.png}}
  \probefig{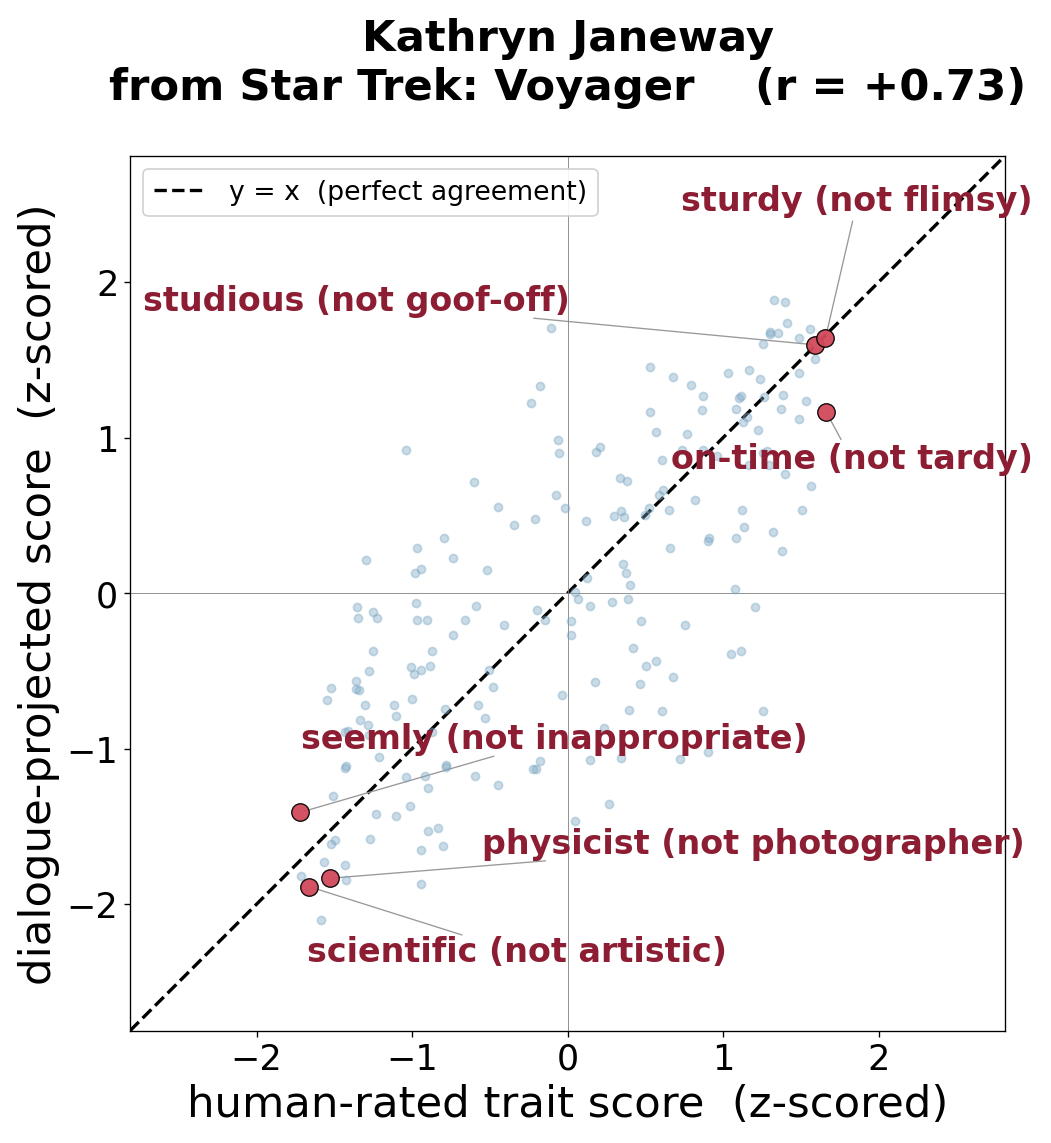}\hfill
  \probefig{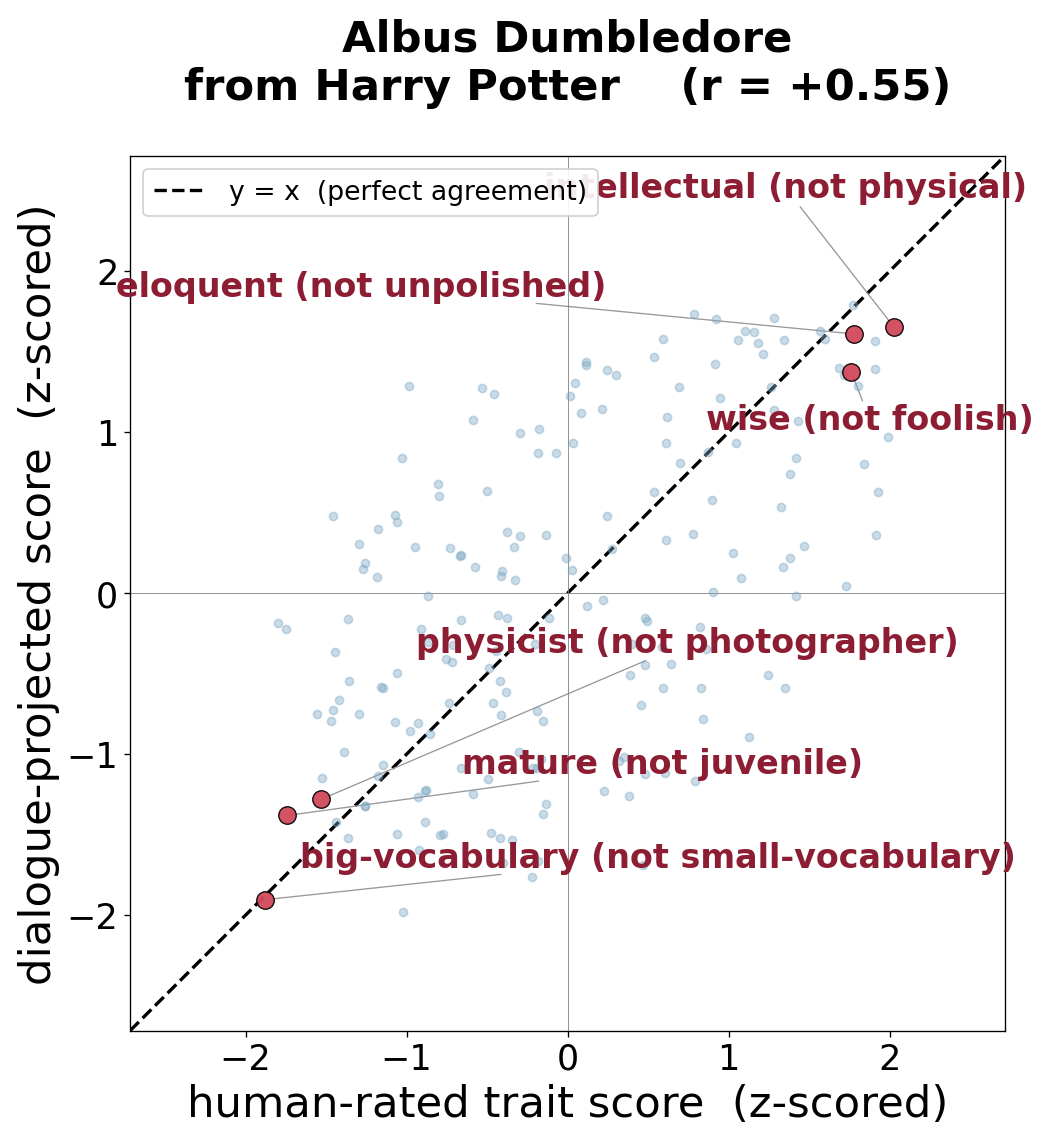}\\[2pt]
  \probefig{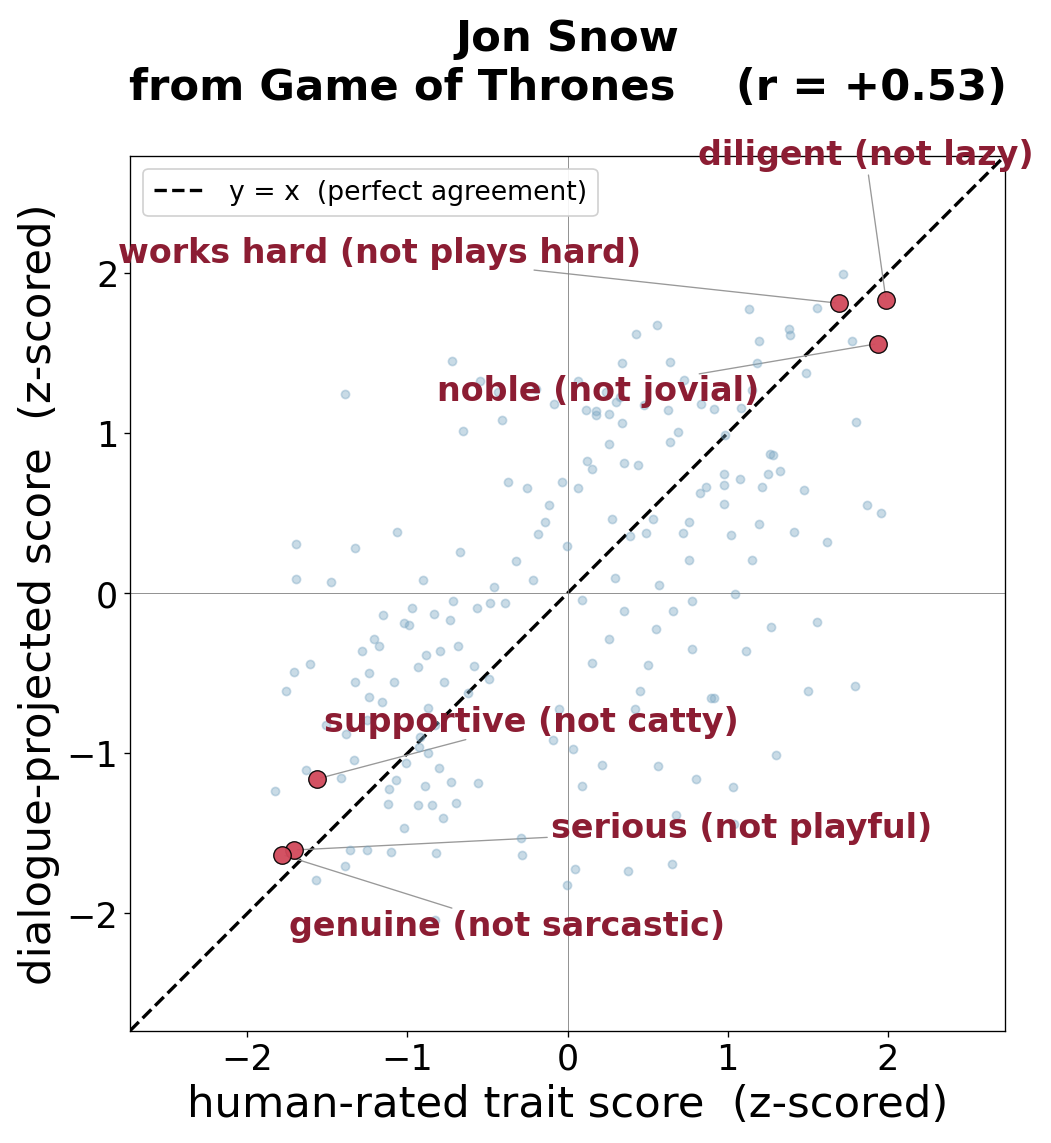}\hfill
  \probefig{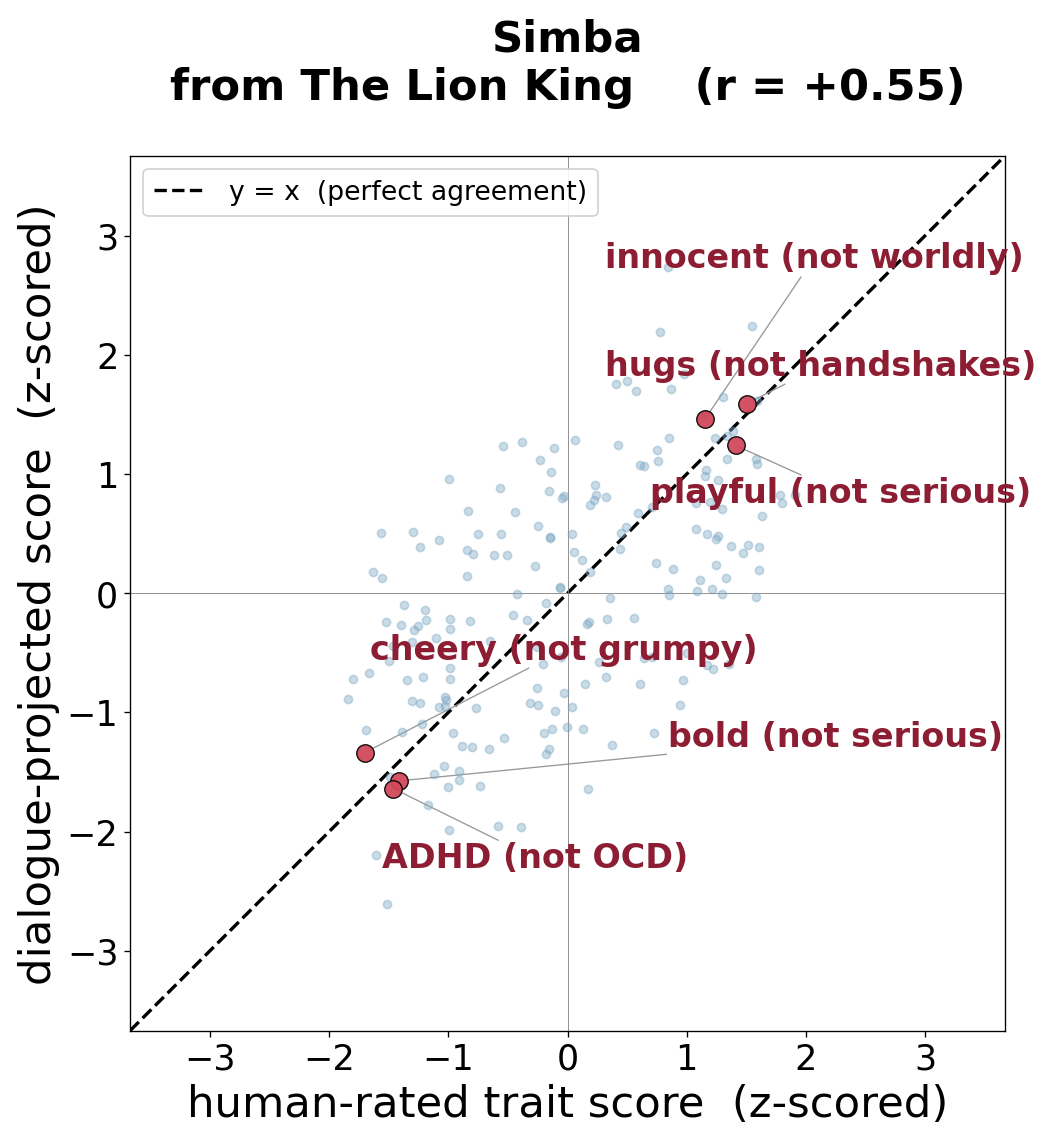}\\[2pt]
  \probefig{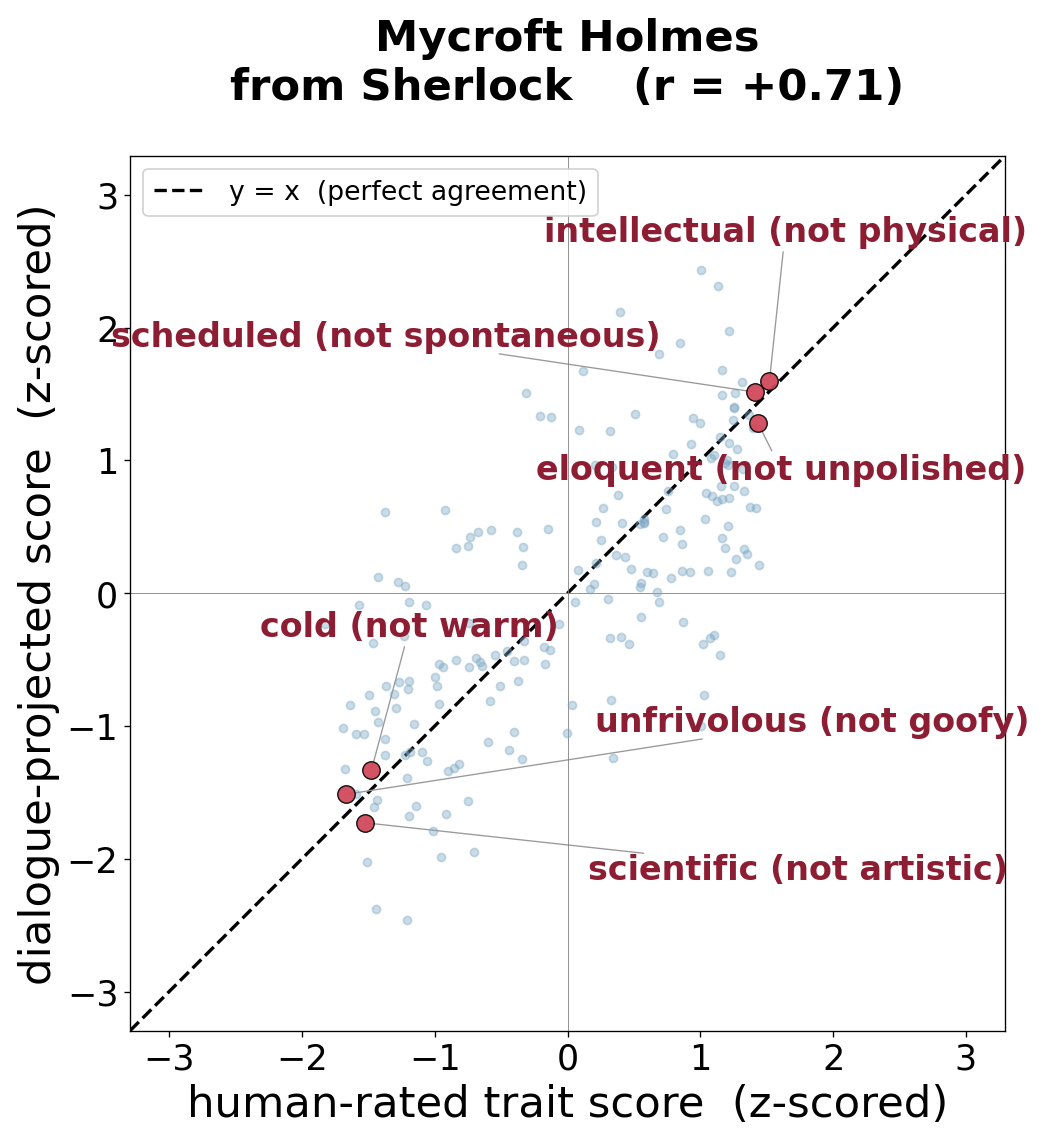}\hfill
  \probefig{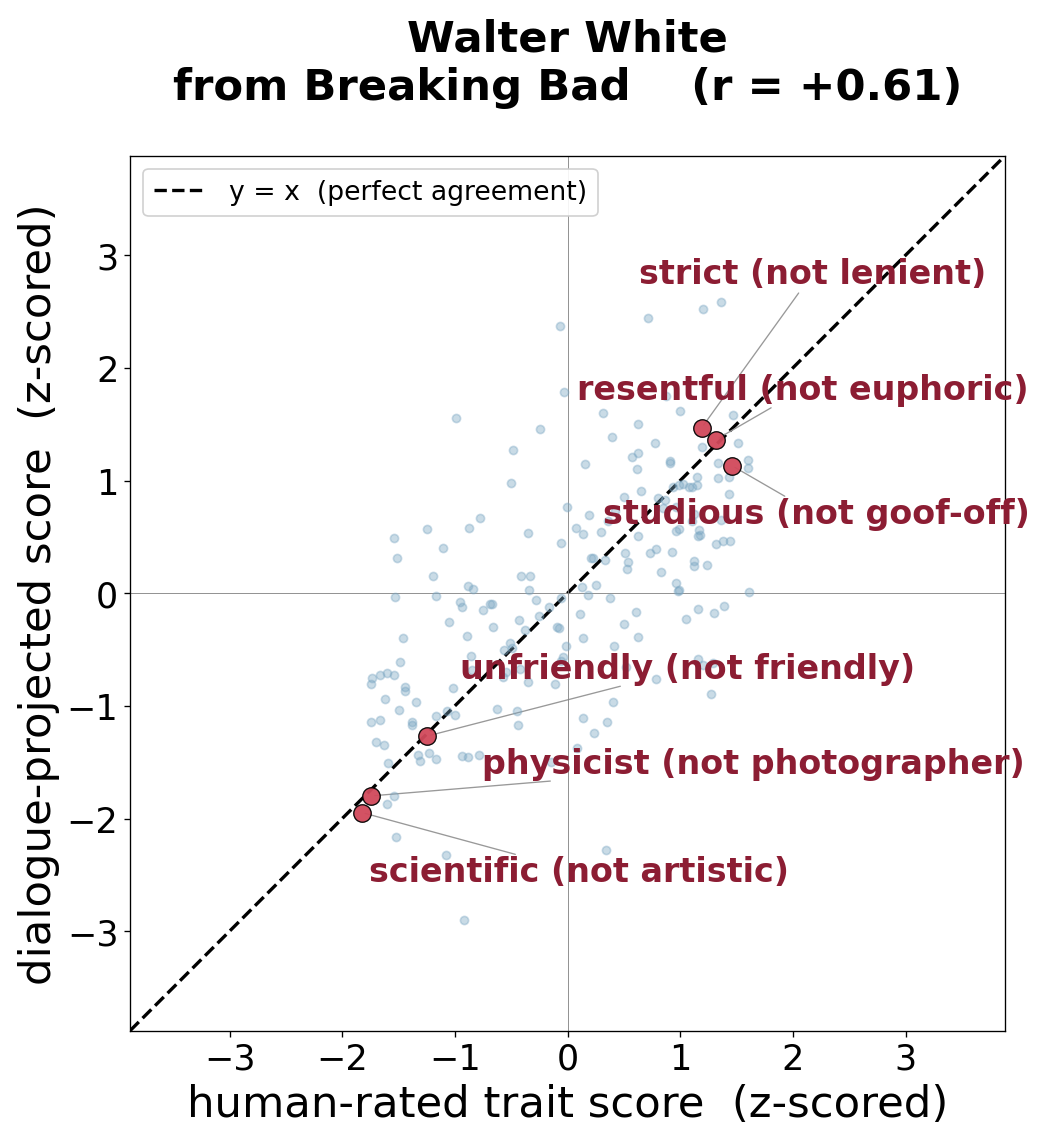}\\[2pt]
  
  \hspace{0.45\textwidth}
  \caption{Dialogue-projected versus human-rated trait scores for eleven held-out characters (cf.\ Figure~\ref{fig:probe_main}). Each point is one of the 195 aligned trait axes, standardised per character; the dashed line is $y=x$. Pearson $r$ (aligned subset) is in each panel title.}
  \label{fig:probe_appendix}
\end{figure*}

\begin{figure}[t]
  \centering
  \includegraphics[width=0.85\linewidth]{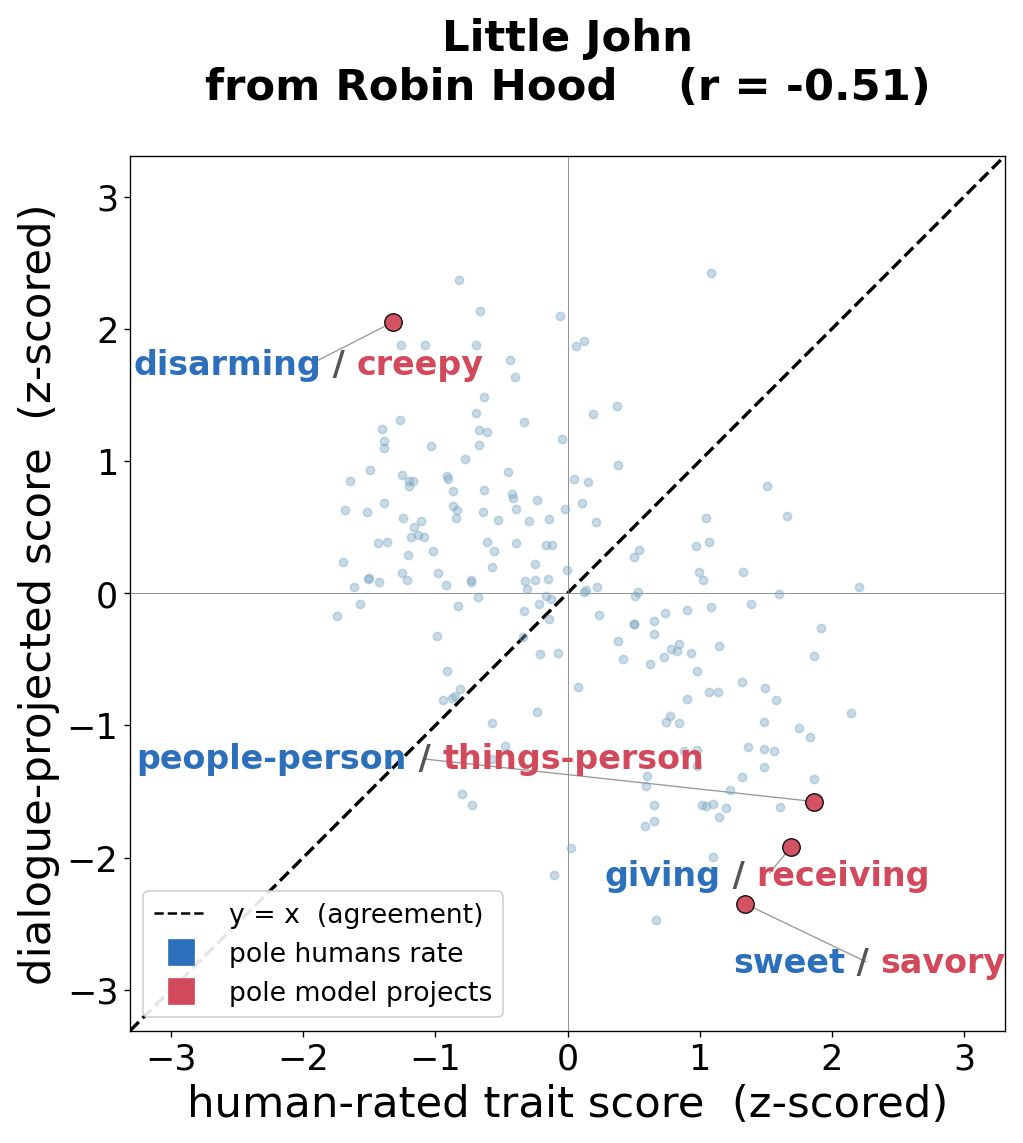}\\[3pt]
  \includegraphics[width=0.85\linewidth]{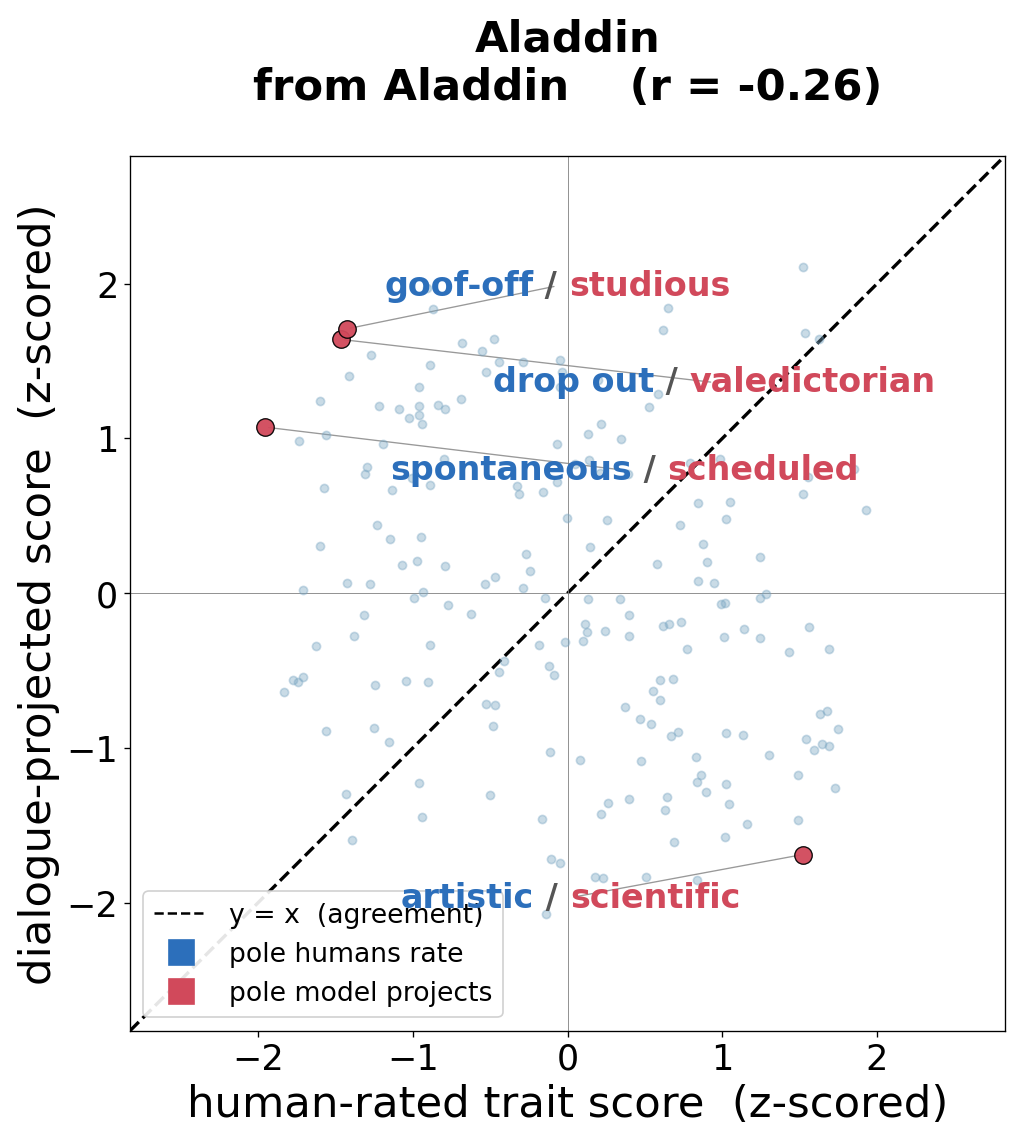}\\[3pt]
  \includegraphics[width=0.85\linewidth]{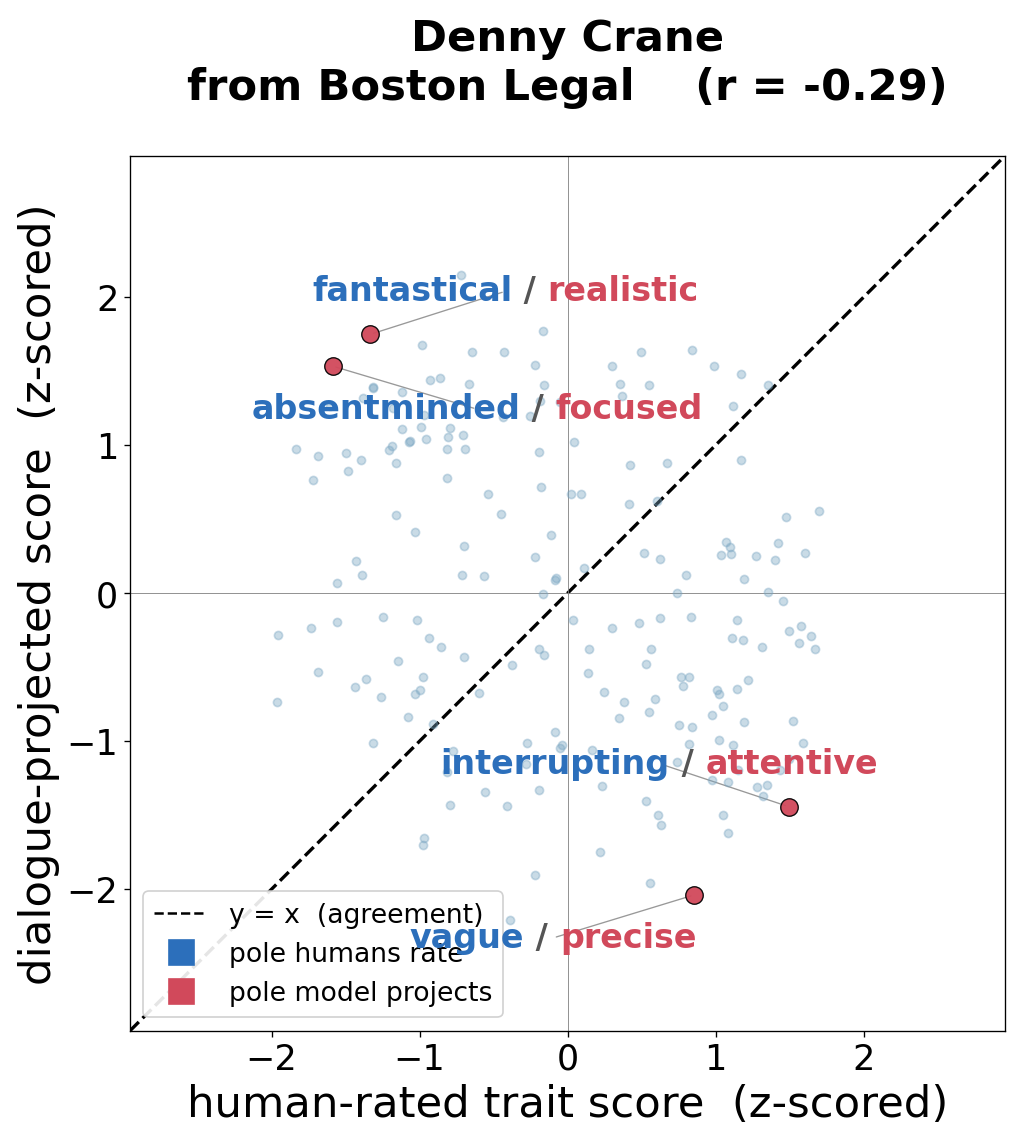}
  \caption{Three meaningful failure examples ($r < 0$). Blue is the pole humans rate, red the pole the dialogue projects. }
  \label{fig:probe_failures_app}
\end{figure}

Figure~\ref{fig:probe_hist} gives the full distribution of per-character agreement over the 256 held-out characters. Figure~\ref{fig:probe_appendix} shows the dialogue-projection probe (Section~\ref{sec:probe}) for more examples, spanning heroes and villains across distinct fictional worlds and the range $r = 0.50$ to $0.73$ on the 195 aligned trait pairs. Each point is one trait axis, standardised per character; the dashed line is $y = x$. Highlighted points are aligned, distinctive traits, labelled with the pole the character is rated toward and its opposite. Figure~\ref{fig:probe_failures_app} shows the three meaningful failures discussed in Section~\ref{sec:probe}, with a two-colour convention: blue is the pole humans rate, red the pole the dialogue projects.
\subsection{Full-Trait Evaluation and Null Baseline}
\label{app:probe_full}

For each of the 256 held-out characters, we mean-pool Qwen 2.5-7B-Instruct's layer-20 hidden states over the recovered Wikiquote input tokens and compute Pearson correlation between the resulting representation and each trait direction. We compare this model profile with the character's complete SWCPQ profile using Pearson correlation. The 256-character cohort has no identifier overlap with the 877 unique characters used as anchors for the final 385 directions.
\begin{table}[t]

\centering
\small
\begin{tabular}{lcccc}
\toprule
Traits & Median $r$ & 95\% CI & Positive & $p_{\rm perm}$ \\
\midrule
195 aligned & 0.403 & [0.358, 0.445] & 238/256 & $<10^{-4}$ \\
All 385 & 0.328 & [0.311, 0.359] & 243/256 & $<10^{-4}$ \\
\bottomrule
\end{tabular}
\caption{Agreement between dialogue-projected and human-rated profiles. Intervals are percentile bootstrap intervals over characters.}
\label{tab:probe_full}
\end{table}
The 95\% intervals use 10,000 character resamples with replacement (seed 20260714). For the null, each draw applies one shared random permutation of the human trait axes to all 256 characters and recomputes the median per-character correlation. No one of 10,000 null medians reaches the observed value for either cohort. Character rankings are highly stable between the full and aligned evaluations (Spearman $\rho=0.966$, $n=256$). These analyses test representational transfer to held-out dialogue, not causal use of the directions during generation.

\end{document}